# Behaviour description database for AVs in Singapore

**CETRAN Team:**

Ana Isabel Garcia Guerra

Teng Sung Shiuan

Paul Hibbard

Yap Jing Yew

Yeo Teck Beng

(30 April 2023)

**TABLE OF CONTENT**

**Disclaimer**

**This research in this report is supported by the National Research Foundation, Singapore, and Land Transport Authority under Urban Mobility Grand Challenge (UMGC-L010).**

# 1. Motivation

The assessments of Autonomous Vehicles (AVs) in the Singapore Milestone framework currently utilises the assessment based on the theory of driving in Singapore that was designed for human drivers based on the highway code. The highway code has some characteristics which make it difficult to be used for assessment of autonomous vehicles, some of these are:

- The testing of human drivers is based on experience of the examiners, and they are assessing that a driver can demonstrate anticipating risk and avoiding it in dynamic traffic instead of reacting to a prescribed set of standardised tests. On the other hand, a key approach applied in autonomous vehicles' scenario-based testing is utilising a set of test scenarios and clear pass or fail criteria. The difference in the test approach is that a human driver is expected to learn and improve in the future after assessment, whereas an AV is expected to perform consistently at its current software configuration level.
- Differences in viewpoint, experience and backgrounds also cause a significant gap in today's test assessments and developers are frustrated with the lack of guidance available to develop solutions for Singapore roads and the test criteria they need to meet to allow their solutions to be deployed.
- Current road traffic act is deliberately written to be non-specific and non-prescriptive in many aspects to ensure Traffic Police assessors have the ability to apply practicality and ambiguity to capture the complexity of day-to-day traffic. This is very different from technical specifications expected by engineers of AV developers, required to develop and verify design logic. However, the level of detail required could be too specific that will not apply to all conditions in everyday dynamic traffic.
- The Highway Code is written for humans and what humans can observe in traffic. It does not consider the state of technology and the differences in terms of information available. AVs sometimes have better observation of surrounding traffic situation of the vehicle compared to a human. Some of the guidance for AVs can be updated to consider AV capability for influencing assessment metrics to be different from current human-driven rules like a following distance to cars ahead. An AV could have the metric to be able to stop with a 2metre gap, different from a 2 second following rule of human driving.

To address some of these ambiguities, investigation into missing guidelines will be carried out within the scope of this report. This effort requires consolidation of current guidelines in Singapore's Standards for Autonomous Vehicles: Technical Reference 68 Part 1 (TR68-1) [1] on basic AV behaviour and current human driving rules from Basic Theory of Driving (BTD) [2] and Final Theory of Driving (FTD) [3] for Singapore roads. It will allow an initial identification of missing guidelines for AV behaviour on roads; however, it is difficult to identify conflicting rules or gaps in guidance without going into identified traffic situations. Identified situations for analysis will be chosen from Centre of Excellence for Testing & Research of Autonomous Vehicle (CETRAN) milestone testing experience for investigation. Research collaboration with an AV developer helped identify some situations and difficulty in determining expected behaviour utilizing existing guidelines. Further analysis of these identified situations will allow for detailed evaluation of conflicting rules or gaps in guidance. Observing safety is priority when developing AV systems for trials on public roads. However, there are situations where observing safe behaviour would result in blocking of traffic flow. TR68-1 has listed two directives in this order, with the prime directive of safety and secondary to maintain traffic flow [1]. This is echoed by Mobileye that considerations for safety may need to be sometimes weighed with the secondary of maintaining traffic flow [4]. One such situation would be an AV operating in a residential area, it could be safest to be wary of pedestrians rushing onto the road and drive very slow [4]. This however impedes traffic flow, hence considerations on balancing safety with traffic flow should be considered.

Development of suggested additional behaviours for AV on Singapore roads should fall within the following boundary conditions:

- Safety operator behaviour is out of the scope of this study.
- AV capability assumed to be Society of Automotive Engineers (SAE) J3016 Level 4/5 [5]
- Mixed traffic environment
- Take reference from existing rules for conventional vehicles.
- Focus on behaviour and not the vehicle capability (perception, decision, etc.)
- Ethical decisions are out of scope i.e. which decision causes less harm
- Study will not cover imminent accidents

An outcome will be to propose additional behaviour characteristics and guidelines to these situations to close the gap between assessors and developers on expected AV behaviour. These recommendations could improve current guidelines for AV behavioural assessment in milestone assessment and generally for the local AV ecosystem for urban tropical roads in Singapore. These recommendations could also serve as inputs for future TR 68-1 revisions where a sample set of reference situations can help to define clearer expectations or requirements for AV behaviour in those situations. This will help Singapore push forward in better definition of the expected AV behaviour for AV systems.

## 2. Approach

### 2.1 Introduction

CETRAN's approach on suggesting additional behaviours for AV on Singapore roads is two folds in creating an AV behaviour database. First, a database consisting of consolidation of existing guidelines is carried out to understand current TR68-1 that was formed for AVs in Singapore and checking with road rules for human-driving such as the BTD and FTD. The second aspect of this behaviour database is the analysis of selected traffic situations and the corresponding additional recommended guidelines. The analysis is done to identify the gaps in guidance or conflicting rules and propose additional guidelines to complement current guidelines.

### 2.2 Consolidation of existing guidelines

The existing guidelines from Singapore's highway code is structured for human drivers learning to drive in Singapore. TR68-1 was then additionally formed to "sets out fundamental behaviours AV exhibits while driving on public roads in order to co-exist safely with entities on the roads such as other vehicles, cyclists and pedestrians" [1]. Fundamental behaviours stated in TR68-1 aims to provide the key directives for automated driving, to apply to AVs for basic interactions "anticipated in the context of Singapore's public roads" [1]. The consolidation of these existing guidelines was done to allow for an efficient lookup of expected behaviours for both assessors and developers due to traffic situation. It can be sub-categorised into 1. General driving behaviours: including lane discipline, speeds, and acceleration; 2. Infrastructure: traffic light junctions, zebra crossing etc; 3. Driving actions: Overtaking, lane change etc.

With experience of assessing autonomous vehicles for trials on public roads, CETRAN has noticed that the look-up for relevant information on expected behaviours within TR68-1 and BTD, FTD requires additional effort. This effort of consolidating existing guidelines in a classification structure would allow for simpler reference. This process allows the team to identify expected behaviours regarding the manoeuvres or at traffic infrastructure in a consolidated manner across TR68-1, BTD and FTD, and identify potential missing or conflicting guidance within the category. This forms the first part of the behaviour database from existing guidelines.

### 2.3 Analysis of identified situations.

Identified situations from CETRAN's experience of assessing AVs for trials on Singapore roads have highlighted some situations where ambiguity of expected behaviour occurred. Through such traffic situations, CETRAN can then identify in a specific manner where guidelines can be improved. Missing guidance could be identified in these situations when analysing the expected behaviour, and conflicting rules also raised from the traffic situation that could be a combination of road infrastructure and intended manoeuvres. Some parameters are guidelines on following distance to incorporate AV detection capability, instead of 2 seconds guidance for human drivers. Others include recommending clearances to other road users in the case where nominal clearances could not be respected, creating some allowances that are typically seen in everyday traffic situations. Some of the recommended guidelines from discussions were then tested on the CETRAN track to check on the practicality of the recommendations. These form the second part of the behaviour database of additional recommended guidelines with the selected traffic situations analysed.

## 2.4 Collaboration with STEAS

A research collaboration with an AV developer in Singapore was established to help CETRAN get insights regarding recommended guidelines in more details when analysing certain identified situations. This was established with Singapore Technologies Engineering Autonomous Solutions (STEAS) who have experience in developing Class 4 vehicles which is the same class such as public buses. This presents another opportunity to engage in guidelines for bigger vehicle sizes when recommending generic guidelines that should apply to all vehicle sizes on public roads in Singapore. Through weekly discussions, STEAS shared situations where they faced challenges with current guidelines and, further discussions were held to develop recommended guidelines. STEAS provided use of their AV bus to undergo trials at CETRAN test track to test out the recommended guidelines which provides valuable insight into the practicality of recommended guidelines to also be applicable to Class 4 vehicles.

## 2.5 Workshop with AV developers

Industry feedback from AV developers on current guidelines is also important to understand their interpretation of existing guidelines and understand their challenge when developing AV for trials on Singapore roads. Their experience from previous assessments and developing for trials provide insight into the practicality of existing guidelines from highway code for human drivers. Furthermore, they could have experience of encounters of traffic situations that are difficult to resolve for an AV with conflicting rules – which humans naturally tend to resolve without much deliberation highlights some issues of applying current existing guidelines for AVs without additional consideration.

# 3. Consolidation of Existing Singapore Guidelines

TR68-1 was established to support the development of AV technology and deployment in Singapore. This guideline presents a list of basic behaviours AVs should display while driving on roads to co-exist safely with other road users and to conform to Singapore traffic rules & regulations.

The Singapore Highway Code – including BTD and FTD is used for teaching human drivers the rules and guidelines of driving in Singapore from the Traffic Police. The driver is trained to be responsible for all driving tasks within Singapore environment. The driver licence provides a basis with the expectation that learning continues with the driving experience after the authorisation.

TR68-1 highlights relevant rules and guidelines from Singapore Highway Code that was meant for human drivers, selecting relevant parameters for AVs in some instances that differ from human drivers. This was first used by the CETRAN team as the effort to distil the rules for AVs from human drivers in Singapore has been initially carried out. TR68-1 has already referenced relevant rules for AVs based on FTD and BTD. From CETRAN's testing experience, additional guidelines from within BTD and FTD were found to be applicable for AVs behaviour specification and have been included in the consolidation. As mentioned earlier, this consolidation of existing Singapore guidelines forms the first part of the behaviour database.

## 3.1 Classification Structure

As mentioned earlier in this project's approach, CETRAN has noticed from experience of AV assessment that substantial effort is required to examine expected behaviours from various sources TR68-1, BTD and FTD. A consolidation of the expected behaviour from these sources allows a simpler and efficient lookup of expected behaviours and allow assessors and developers to identify expected behaviours based on manoeuvres or at traffic situations.

The consolidation is grouped into the 3 categories below with the mindset that AV stakeholders can easily look up to the associated behaviours on route. The first general driving behaviours would apply to overall driving – for smoothness and comfort of the ride. Secondly would be directly on infrastructure requirements – to distil the applicable guidelines for an AV such as when approaching a traffic intersection or travelling from a major to a minor road on its route. Thirdly, it relates to expected behaviours of AV actions due to traffic situation, such as a lane change manoeuvre. From the analysis of identified situations, it has proved that this classification structure has been useful to identify relevant guidelines.

The classification structure for CETRAN's consolidated driving behaviours for AVs on Singapore roads are in the following:

1. General driving behaviours: including lane discipline, ride comfort

2. Infrastructure: traffic light junctions, zebra crossing etc.

3. Driving actions or manoeuvre related actions: Overtaking, lane change etc.

Table 1: Table showing structure of current guidelines.

| TR 68-1 | BTD |
|---|---|
| 1) Safe Distances and Speeds | 1) Traffic Signs and Signals |
| 2) Traffic situations and manoeuvres | 2) Traffic Rules and Regulations |
|  | 3) Code of Conduct on the Road |

**General driving behaviours:**

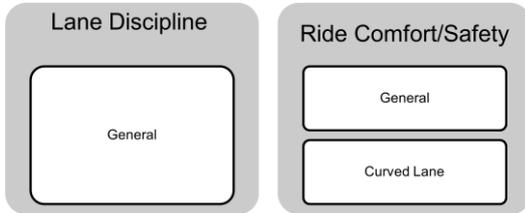

**Infrastructure:**

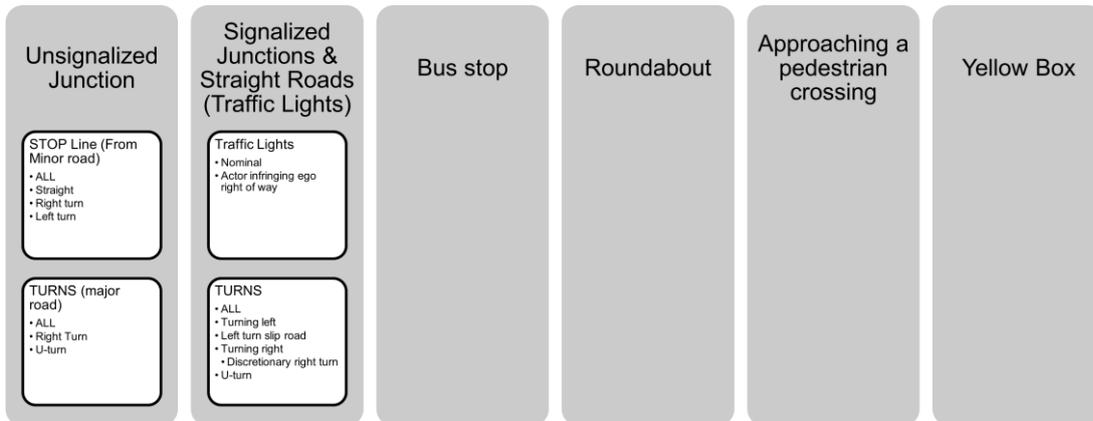

**Driving actions/ Manoeuvre related actions:**

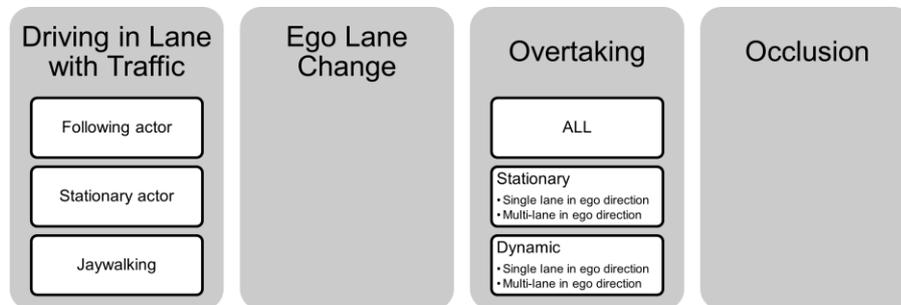

## 3.2 Consolidated AV Behaviour Database

### 3.2.1 General driving behaviours

#### 3.2.1.1 Lane Discipline

Table 2: Consolidated guidelines from TR68-1, BTD & FTD on Lane Discipline

| General | Parameters |
|---|---|
| **TR68-1:** 7.3 An AV shall adhere to traffic signs and road markings regardless of the information source used by it, such as live detection and/or map database. | |
| **TR68-1:** 7.4a) The AV shall obey directions indicated by arrows marked in lanes **(BTD64).** | |
| **TR68-1:** 7.4b) The AV shall keep to the left of a two-lane carriageway, except when overtaking **(BTD67, BTD160a, FTD310)**. | |
| **TR68-1:** 7.4d) The AV shall keep within its lane unless it is performing a lane change or overtaking manoeuvre **(FTD311, 160b).** | |
| **BTD 160c),FTD312** Do not weave in and out of traffic lanes as it would cause confusion and danger to others | |
| **FTD108**: Always keep well to the left when driving along two-way streets or dual-carriageways, unless your path of travel is obstructed by road works, parked vehicles, etc. In such circumstance, you may cross the centre line or move to the lane on your right. On doing so, take care to ensure that your intended path is safe and clear before you move to the right. | |

#### 3.2.1.2 Ride Comfort/Safety

##### 3.2.1.2.1. General

Table 3: Consolidated guidelines from TR68-1, BTD & FTD on Ride Comfort and Safety - General

| General | Parameters |
|---|---|
| **TR68-1:** 6.1.a) Unless otherwise stated, the speed limit of all roads in Singapore is 50 km/h. Do not exceed the speed limit of your vehicle or the road, whichever is lower**. (BTD78)** | 50km/h |
| **TR68-1:** 6.1.b) Adhere to posted variable/temporary speed limits and speed warning signs **(BTD207, BTD245, BTD79)** | |
| **TR68-1:** 6.1.d) Adjust speed to account for advisory speed signs, designated calmed areas (school, silver) and road scenery | |
| **TR68-1:** 6.1.e) Constantly and smoothly adjust speed for road alignment and surface conditions. The selected speed should allow for good traction to be maintained, accounting for path curvature, road conditions, and potential longitudinal acceleration. | |
| **BTD179/FTD177:** The basic rules to follow when driving in bad weather conditions are:<br>(a) Reduce your speed so that you can manoeuvre safely.<br>(b) Do not out-drive the actual distance that you can see clearly.<br>(c) Switch on your headlights so that you can see more clearly and be more visible.<br>(d) Turn on the demister to clear the mist on the windscreen.<br>(e) Move to a safe spot and stop at the side of the road with your hazard lights turned on if you cannot see clearly. Proceed when conditions are favourable.<br>(f) Use a faster front windscreen wiper speed when driving in heavy rain. | |
| **BTD180/FTD178:** On a wet road, the stopping distance of a vehicle will increase to about twice the distance of that on a dry road. This is because there is less friction between the tyres and the wet road surface. On a wet road, stepping hard on the brake pedal (locking the wheels) will cause the vehicle to skid or spin. If you lock the wheels accidentally, quickly release the brake pedal and apply the intermittent (ON/OFF) braking technique until the vehicle comes to a stop. Thus, on a wet road, it is important that you drive at a slower speed to avoid such dangerous circumstances | |
| **BTD181/FTD179:** On a rainy day, a thin layer of water forms on the road surface. Even good tyres may not have a good grip on the road. As speed increases, surface water builds up under the vehicle's tyres. When this happens, your vehicle will glide on the surface of the road, and this is known as 'Aquaplaning'. | |

| | Parameters |
|---|---|
| **BTD195/FTD193:** A good driver should know how to read and adjust his/her speed accordingly to suit the road conditions, such as:<br>(a) The width of the road;<br>(b) The kind of road surface he/she is driving on;<br>(c) The contour of the road;<br>(d) The possible danger of hazards ahead. | |
| **FTD199:** Adjust your speed to the traffic and road situation. Situations change as you travel from urban built-up areas to sub-urban and then to rural areas. | |

### 3.2.1.2.2. Curved lane

Table 4: Consolidated guidelines from TR68-1, BTD & FTD on Ride Comfort and Safety on Curved Lanes

| Curved Lanes | Parameters |
|---|---|
| **BTD199/FTD197:** The greater the travelling speed around the curve or the sharper the curve, the more the vehicle will be pushed from its path. You should therefore reduce speed when going round a bend. The diagram on the right shows the appropriate speed and the dangerous speed for each turning radius. | |

## 3.2.2 Traffic Infrastructure

The following section is on the guidelines for driving at infrastructure. Some of these are broken down into intended actions at these infrastructure – such as an AV route having a right turn at a signalized junction.

### 3.2.2.1 Signalized Junctions (Traffic Lights)

#### 3.2.2.1.1. Traffic Lights – Nominal

During the route, an AV reaches the road infrastructure of a signalized junction and encounter traffic light signals at junction. The following are the guidelines.

Table 5: Consolidated guidelines from TR68-1, BTD & FTD on Nominal behaviour at traffic light infrastructure

| GENERAL- Traffic Lights/ Nominal | Parameters |
|---|---|
| **TR68-1:** 7.2.2) The AV should be capable of detecting and interpreting the traffic signals | |
| **TR68-1:** 7.2.3) The AV should be capable of identifying which traffic signal aspect groups are controlling the traffic stream that corresponds to an AV's path through a signal-controlled area. This is termed as governing signal group. | |
| **TR68-1:** 7.2.4) The AV should then adhere to directions of the governing signal group with the meanings of each signal being defined in **BTD 57**. When state of governing signal group changes from green to amber:<br>7.2.4a) AV should try to stop before the stop line if it is safe to do so<br>7.2.4b) AV should continue passing through the stop line in a safe and appropriate manner if stopping would require aggressive braking | |
| **TR68-1:** 7.2.5) When traffic signal turns from green to amber, an AV should stop at the stop line as described in 6.4, if this is possible with the comfortable stopping distance described in 6.2. If that is not possible, then the AV should continue to approach and cross the intersection at a velocity at or below the applicable speed limit. | |
| **TR68-1:** 7.2.6d The AV should adjust its speed on approach to the junction (**FTD118**) in order to<br>i) be at a suitable speed to safely stop for crossing pedestrians, if applicable<br>ii) be able to comfortably stop before the stop line should the governing signal group change to amber. | |

| Traffic Lights / continued | Parameters |
|---|---|
| **TR68-1:** 7.2.6e) The AV should not proceed past the stop line if there is no space for exiting the junction (e.g. due to queued vehicles (**FTD120**) | |
| **TR68-1:** 7.10.3 c) When going straight, avoid using the shared lane so as not to obstruct vehicles turning right when the traffic signal changes to red for vehicles going straight (**FTD117**). | |

### 3.2.2.1.2. *Traffic Lights- Actor Infringing right of way*

Table 6: Consolidated guidelines from TR68-1, BTD & FTD at Traffic Lights – where other actors infringe right of way.

| Traffic Lights/ Actor Infringing right of way | Parameters |
|---|---|
| **TR68-1:** 7.2.6b) Do not strictly follow a right of way when governing signals are green. Other vehicles may run through the opposing red light (**BTD 204b.6**) including emergency vehicles; right-turning vehicles may turn into the opposing vehicle's path. | |
| **TR68-1:** 7.2.6c) Give way to pedestrians crossing the road (**BTD145**) <br> **BTD145**) When approaching a pedestrian crossing, ALWAYS – <br> (a) be ready to slow down or stop so as to give way to pedestrians; <br> (b) signal to other drivers your intention to slow down or stop; <br> (c) allow yourself more time to stop when the road is wet | |
| **BTD 146**) At a pedestrian crossing controlled by traffic signals or by a policeman, give way to pedestrians who are still crossing even when the signal allows vehicles to move. | |

### 3.2.2.1.3. *TURNS – ALL (General)*

Table 7: Consolidated guidelines from TR68-1, BTD & FTD for making a turn in general at a traffic light junction

| TURNS – ALL (General) | Parameters |
|---|---|
| **TR68-1:** 7.2.6d iii) (FTD 118) be at a suitable speed in readiness for any planned manoeuvres at the junction (eg. Turning left or right) <br> **FTD118:** Do not accelerate when you are approaching a signalized junction. Always be prepared for the signal light to change by reducing your speed. This will give you time to decelerate evenly when the traffic light changes from green to amber. | |
| **TR68-1:** 7.6 f)ii) An AV shall always overtake to the right, except when it is turning left at a junction ahead | |
| **TR68-1:** 7.6 g)ii and iii An AV shall not perform an overtaking manoeuvre at, or when approaching ii. A road junction or and iii. A corner or bend | |
| **TR68-1:** 7.10.2 a) When turning left or right at a junction, signalised or non-signalised, an AV shall keep to its lane corresponding to its lane before turning. | |
| **TR68-1:** 7.10.2 b) An AV shall enter the lane that permits the intended direction of turn at a junction or a slip road as indicated by road markings. | |
| **TR68-1:** 7.10.2 c) An AV shall signal its intention at least 3 seconds before entering the lane which permits the intended direction of turn at a junction or a slip road. | 3s |

| |
|---|
| **BTD93:** It is an offence for any driver to drive his/her vehicle into a junction marked with a yellow box and cause obstruction even if the lights are in his/her favour. However, there are certain situations in which a driver or rider will not be penalised for entering the yellow-box junction.<br>• **BTD93(a) Situation One**: The first is when the turning vehicles in a box-junction do not block other vehicles. Only drivers of turning vehicles A, B and C and NOT those marked X may enter the yellow box when their exit lanes from it are blocked by other vehicles. Should the traffic lights change to red, vehicles A, B and C would not cause any obstruction to other vehicles. However, vehicles marked X would be obstructing traffic should the lights change and the drivers will be penalised, as such they should wait behind the stop-line at 'Y'.<br>• **BTD93b) Situation Two:** The second situation is when motorists are waiting in the yellow box while trying to turn right in the face of oncoming traffic. Drivers of vehicles marked A, B and C can remain in the box until a suitable opportunity arises for them to complete their turns. Should the lights turn red against them, they should proceed to complete their turn and clear the yellow box junction.<br>• **BTD93c) Situation Three**: The third situation is when motorists are waiting in a yellow box junction while making left or right turns because of pedestrians crossing the road. The diagram shows that the drivers of vehicles A, B, C, D and E may enter the yellow box even when they are prevented from executing their turns by pedestrians using the crossings. |

### 3.2.2.1.4. <u>TURNS – Turning left</u>

Table 8: Consolidated guidelines from TR68-1, BTD & FTD on turning left specifically at a traffic light junction.

| TURNS- Turning Left | Parameters |
|---|---|
| **TR68-1:** 7.10.3 ) An AV shall keep its signal until the turn is completed. When turning left at a signalised junction, an AV shall first stop to allow pedestrians, cyclists and PMD riders to cross the road, before proceeding to complete the turn. | |
| **BTT33:** (Mandatory Signs) <u>Left Turn on Red</u><br>•Stop at the red light<br>•Give way to pedestrians crossing at the junction<br>•Give way to traffic approaching from the right<br>•Proceed only when the way is clear and it is safe to do so | |
| **BTT45:** (EMAS Signs) No Left Turn (Arterial Road –Left turn lane(s) closed ahead) | |
| **FTD 113 (1)**: At 'a' Check your front and rear for vehicles and other road users.<br>At 'b' Signal your intention early.<br>At 'c' Check your left blind spot and if it is safe, keep left.<br>At 'd' Reduce your speed.<br>At 'e' Check if the traffic light allows you to proceed. If not, stop behind the stop line.<br>At 'f' Look right and ahead to confirm that it is safe to proceed. Keep a look-out for two-wheeled vehicles before you make the turn. Always give way to pedestrians crossing at the junction.<br>At 'g' Turn into the nearest lane. After turning left, increase your speed accordingly | |

### 3.2.2.1.5. <u>TURNS – left turn slip road</u>

Table 9: Consolidated guidelines from TR68-1, BTD & FTD for specifically entering a left turn slip road at a traffic light junction

| TURNS-Left Turn Slip Road | Parameters |
|---|---|
| **TR68-1:** 7.10.2) An AV shall comply with the "Give Way" rule at non-signalised junctions **(BTD81) BTD81:** At an uncontrolled junction where there are no traffic lights, "GIVE WAY" sign or "STOP" sign: | |
| **TR68-1:** 7.10.2 a) When turning left or right at a junction, signalised or non-signalised, an AV shall keep to its lane corresponding to its lane before turning. | |
| **TR68-1:** 7.10.2 b) An AV shall enter the lane that permits the intended direction of turn at a junction or a slip road as indicated by road markings. | |
| **TR68-1:** 7.10.2 c) An AV shall signal its intention at least 3 seconds before entering the lane which permits the intended direction of turn at a junction or a slip road. | Signal-3seconds |

### 3.2.2.1.6. TURNS – turning right

Table 10: Consolidated guidelines from TR68-1, BTD & FTD for specifically turning right at a traffic light junction

| **TURNS- Turning Right** | **Parameters** |
|---|---|
| **TR68-1:** 7.10.3 b) When turning right going straight, into a shared lane for turning right and going straight, signal early so that straight-going vehicles from behind can take an alternate lane to proceed **(FTD116).** <br> **FTD 116:** When turning right into a 'shared' lane for turning right and going straight, signal early so that straight-going vehicles from behind can take an alternate lane to proceed. | |
| **BTD84)** When turning right into a two-way street, turn into the lane just left of the centre line of the road you are turning into. | |
| **BTD85)** When turning right into a one-way street, turn into the extreme right lane of the road you are turning into. | |
| **FTD113 (3)** <br> At 'a' Check your front and rear for vehicles and other road users. <br> At 'b' Signal your intention early. <br> At 'c' Check your right blind spot and if it is safe, take up the appropriate lane as indicated by the road markings. <br> At 'd' Reduce your speed. <br> At 'e' Check if the traffic light allows you to proceed. If not, stop behind the stop line. <br> At 'f' Drive slowly and cautiously towards the centre of the junction; Give way to oncoming vehicles; Wait until it is safe to cross or wait for the green arrow signal to appear; Turn swiftly to the correct lane keeping a look-out for pedestrians crossing at the junction. <br> At 'g' Increase your speed accordingly. | |

### 3.2.2.1.7. Turning Right – Discretionary right turn

Table 11: Consolidated guidelines from TR68-1, BTD & FTD for specifically turning right at a traffic light junction, focused on discretionary right turns.

| **TURNS- Turning Right *Discretionary right turn (no right of way)*** | **Parameters** |
|---|---|
| **TR68-1:** 7.10.3 a) When turning right at a signalised junction, an AV shall first stop at **the right turning pocket**. Otherwise, where there is no right turning pocket, an AV shall first stop at a safe distance away from opposing traffic. The AV shall only proceed to make the right turn when there are neither oncoming vehicles travelling in the opposite direction nor pedestrians, cyclists and PMD riders crossing the road on its right. | |
| **BTD 54b)** Vehicles turning right at road junctions should stay within the pocket until it is clear to complete the turn. | |

### 3.2.2.1.8. TURNS- U-turn

Table 12: Consolidated guidelines from TR68-1, BTD & FTD for specifically U-turns at a traffic light junction

| **TURNS- U-turn** | **Parameters** |
|---|---|
| **TR68-1:** 7.8) An AV shall only perform a U turn manoeuvre when all the following is fulfilled. <br> a) There is a U-turn sign in the case of turning at an intersection, junction or opening in the road divider (**BTD 94**) <br> b) The AV path is clear of traffic coming from the opposite direction <br> c) There are no ORUs turning together with an AV on its right | |
| **BTD 94** Do not make a U-turn at any road intersection, junction, or any opening in a road divider except where a U-turn sign is located. | |
| **BTD 95** Always negotiate a U-turn carefully, especially if roadside trees or hedges are obstructing your view of oncoming traffic. As you turn, be alert for any vehicles, especially motorcycles, which may be turning together with you on your right. | |

### 3.2.2.2    Unsignalized Junctions

*During the route, an AV reaches the road infrastructure of an unsignalized junction.*

Table 13: Consolidated guidelines from TR68-1, BTD & FTD for approaching an unsignalized junction.

| GENERAL | Parameters |
|---|---|
| **TR68-1:** 7.6 g)ii) An AV shall not perform an overtaking manoeuvre at, or when approaching a road junction | |

#### 3.2.2.2.1.    STOP Line (from minor road) – ALL

Table 14: Consolidated guidelines from TR68-1, BTD & FTD for approaching a Stop line at an unsignalized junction, from a minor road to a major road.

| From Minor Road - GENERAL | Parameters |
|---|---|
| **TR68-1:** 6.1.h) When merging into major roads from slip lanes, match speed of traffic on the major road for a smooth merge. | |
| **TR68-1:** 6.4) When stopping at a stop line, the AV should achieve a stationary state with a longitudinal distance of its foremost point to the nearest edge of the stop line between 0 m and 1.5 m. | Threshold: 0m-1.5m |
| **TR68-1:** 7.2.1) All stops at stop line should be performed in accordance with the provisions set out in 6.4 | |
| **TR68-1:** 7.10.1) When approaching a non-signalised junction with a major road, an AV shall reduce its speed and give way to traffic on the major road. If there is a STOP sign, an AV shall stop at the stop line with the provisions of 6.4. | |
| **TR68-1:** 7.10.2) An AV shall comply with the "Give Way" rule at non-signalised junctions (BTD81)<br>**BTD81**: At an uncontrolled junction where there are no traffic lights, "GIVE WAY" signor "STOP" sign: | |
| **BTD 80**: When approaching a junction with a major road, slow down gradually and give way to traffic on the major road. Where there is a "STOP" sign, stop before the stop-line. | |

#### 3.2.2.2.2.    STOP Line (from minor road) – Going Straight

Table 15: Consolidated guidelines from TR68-1, BTD & FTD for approaching an unsignalized junction, from a minor road and going straight.

| GENERAL | Parameters |
|---|---|
| **BTD 81a)** If you are going straight across the junction, you must give way to traffic going straight from the right. | |

#### 3.2.2.2.3.    STOP Line (from minor road) – Right Turn

Table 16: Consolidated guidelines from TR68-1, BTD & FTD for approaching a Stop line at an unsignalized junction, from a minor road with the intention to turn right.

| From Minor Road – Turning Right | Parameters |
|---|---|
| **TR68-1:** 6.1.h) When merging into major roads from slip lanes, match speed of traffic on the major road for a smooth merge. | |
| **TR68-1:** 6.4) When stopping at a stop line, the AV should achieve a stationary state with a longitudinal distance of its foremost point to the nearest edge of the stop line between 0 m and 1.5 m. | Threshold: 0m-1.5m |
| **TR68-1:** 7.2.1) All stops at stop line should be performed in accordance with the provisions set out in 6.4 | |
| **TR68-1:** 7.10.1) When approaching a non-signalised junction with a major road, an AV shall reduce its speed and give way to traffic on the major road. If there is a STOP sign, an AV shall stop at the stop line with the provisions of 6.4. | |
| **TR68-1:** 7.10.2) An AV shall comply with the "Give Way" rule at non-signalised junctions (BTD81)<br>**BTD81**: At an uncontrolled junction where there are no traffic lights, "GIVE WAY" sign or "STOP" sign: | |
| **BTD 80**: When approaching a junction with a major road, slow down gradually and give way to traffic on the major road. Where there is a "STOP" sign, stop before the stop-line. | |

### 3.2.2.2.4.  STOP Line (from minor road)  – Left Turn

Table 17: Consolidated guidelines from TR68-1, BTD & FTD for approaching a Stop line at an unsignalized junction, from a minor road with the intention to turn right.

| **From Minor Road – Turning LEFT** | **Parameters** |
|---|---|
| **BTD81c**) If you are turning left, you must give way to traffic going straight from the right. | |
| **TR68-1:** 7.10.2 a) When turning left or right at a junction, signalised or non-signalised, an AV shall keep to its lane corresponding to its lane before turning. | |
| **TR68-1:** 7.10.2 b) An AV shall enter the lane that permits the intended direction of turn at a junction, or a slip road as indicated by road markings. | |
| **TR68-1:** 7.10.2 c) An AV shall signal its intention at least 3 seconds before entering the lane which permits the intended direction of turn at a junction or a slip road. | |
| **FTD 246: When turning**<br>(a) check your mirrors and blind spots;<br>(b) look out for cyclists between you and the kerb;<br>I do not make a sharp turn as you may knock the cyclist down. Slow down and give way to the cyclist if it is not safe to turn. | |

### 3.2.2.2.5.  TURNS from Major Road

Table 18: Consolidated guidelines from TR68-1, BTD & FTD for approaching an unsignalized junction, from a major road for all turns in general.

| **From Major Road –Turns General** | **Parameters** |
|---|---|
| **TR68-1: 7.10.2 a**) When turning left or right at a junction, signalised or non-signalised, an AV shall keep to its lane corresponding to its lane before turning. | |
| **TR68-1:** 7.10.2 b) An AV shall enter the lane that permits the intended direction of turn at a junction, or a slip road as indicated by road markings. | |
| **TR68-1:** 7.10.2 c) An AV shall signal its intention at least 3 seconds before entering the lane which permits the intended direction of turn at a junction or a slip road. | |
| **FTD 246: When turning**<br>(a) check your mirrors and blind spots;<br>(b) look out for cyclists between you and the kerb;<br>I do not make a sharp turn as you may knock the cyclist down. Slow down and give way to the cyclist if it is not safe to turn. | |

### 3.2.2.2.6.  TURNS from Major Road – Right Turn

Table 19: Consolidated guidelines from TR68-1, BTD & FTD for approaching an unsignalized junction, from a major road for making right turns.

| **From Major Road –Turns→ Right turn** | **Parameters** |
|---|---|
| **BTD84**) When turning right into a two-way street, turn into the lane just left of the centre line of the road you are turning into. | |
| **BTD85**) When turning right into a one-way street, turn into the extreme right lane of the road you are turning into. | |

![CETRAN]### 3.2.2.2.7. TURNS from Major Road – U-Turn

Table 20: Consolidated guidelines from TR68-1, BTD & FTD for approaching an unsignalized junction, from a major road for making U- turns.

| From Major Road –Turns→ U-turn | Parameters |
|---|---|
| **TR68-1:** 7.8) An AV shall only perform a U turn manoeuvre when all the following is fulfilled<br>a) There is a U - turn sign in the case of turning at an intersection, junction or opening in the road divider (**BTD 94**)<br>b) The AV path is clear of traffic coming from the opposite direction<br>c) There are no ORUs turning together with an AV on its right<br>**BTD 94** Do not make a U-turn at any road intersection, junction or any opening in a road divider except where a U-turn sign is located. | |
| **BTD 95** Always negotiate a U-turn carefully, especially if roadside trees or hedges are obstructing your view of oncoming traffic. As you turn, be alert for any vehicles, especially motorcycles, which may be turning together with you on your right. | |

### 3.2.2.3 Roundabouts

Table 21: Consolidated guidelines from TR68-1, BTD & FTD for approaching an unsignalized junction, from a major road for making U- turns.

| Roundabout / entering | Parameters |
|---|---|
| **TR68 - 7.7.1** When approaching a roundabout, an AV shall reduce its speed and move to the lane corresponding to the exist it intent to take (**FTD 132 a),b),c), FTD135, BTD 89**) | |
| **TR68 - 7.7.2** the AV shall give clear signals showing the intention to enter or exit the roundabout in good time<br>a) if intending to take the first exit, the AV shall signal left at least **3 seconds** before entering a roundabout<br>b) if not intending to take the first exist, the AV the AV shall signal right at least **3 seconds** before entering a roundabout and then switch to signalling left after passing the exist that is comes before the intended exit<br>c) before entering a roundabout, the AV shall give way to traffic that is inside the roundabout and coming from the right (**BTD 88, FTD 133a, FTD 134**) | Show intention 3 second before |
| **FTD 132 (d)** When approaching a roundabout beware of the speed and positions of the traffic around you. | |
| **FTD 133** When entering a roundabout:<br>(b) Look out for traffic already in the roundabout.<br>(c) Do not enter a roundabout if you can see that your exit is blocked. | |

### 3.2.2.4 Pedestrian Crossing

*Pedestrian crossing is a specified part of a road where pedestrians have right of way to cross.*

Table 22: Consolidated guidelines from TR68-1, BTD & FTD for approaching a pedestrian crossing.

| Approaching a pedestrian crossing | Parameters |
|---|---|
| **TR68-1 - 7.6 g) i)** An AV shall not perform an overtaking manoeuvre at, or when approaching a pedestrian crossing | |
| **TR68-1 7.9.1** Giving right of way.<br>An AV shall give right of way to VRU at pedestrian crossings, including zebra crossing unless they have no intent to cross **[ BTT 143] [FTT233]** | |
| **TR68-1 7.9.2** Approaching a pedestrian crossing<br>When approaching a pedestrian crossing, which includes zebra crossing, the AV should adjust its speed such that it can reasonably stop **[ BTT 51d/e]** | |
| **TR68-1 7.9.3** Zebra crossing zone- Definition (See Figure 1 below) | |
| **TR68-1 7.9.4** Stopping at Zebra crossing | When: |

Copyright©2023 Nanyang Technological University

| | |
|---|---|
| AV shall stop outside the zebra crossing zone when VRU are present inside the zebra crossing zone with one or more of the following attributes:<br>a) On the road surface less than 4 m away laterally from the AV footprint<br>b) Moving towards the AV's path and is within 7m laterally from the AV footprint<br>c) Has displayed the intent to cross the AV's path and is within 7m laterally from the AV's footprint.<br>AV shall stop outside the zebra crossing zone at the stop line in accordance with TR 68 -1 6.4 or stopping at stop lines.<br>When stopping at stop lines the AV should achieve a stationary state with a longitudinal distance of its foremost point to the nearest edge of the stop line between 0 and 1,5 m or when no stop line is present 3m before the zebra crossing stripes | Lateral distance of 4m or 7m from the AV footprint<br>How: Longitudinal distance between 0 and 1.5 m of the stop line |
| **TR 68-1 7.9.5** Assumption of pedestrians' intent occupying zebra crossing zone **[BTT 145 a] [FTT236 A]**<br>VRU who are occupied the zebra crossing zone and are not moving away from the path of the AV shall also be assumed to have intent to cross at least until the AV comes to a complete stop at the pedestrian crossing | |
| **TR 68-1 7.9.6** AV's speed limit when traversing the zebra crossing zone<br>The AV shall traverse the zebra crossing zone at the speed of not exceeding 30km/h when the zebra crossing zone is occupied by a VRU who has not displayed intent to cross within the pass of 5 seconds | 30 km/h speed max when VRU does not cross after 5 s |
| **BTD 145 a** When approaching a pedestrian crossing, ALWAYS allow yourself more time to stop when the road is wet | |
| **BTD 147** Stop, when signalled to do so by a school patrol warden showing a "STOP-Children" sign | |

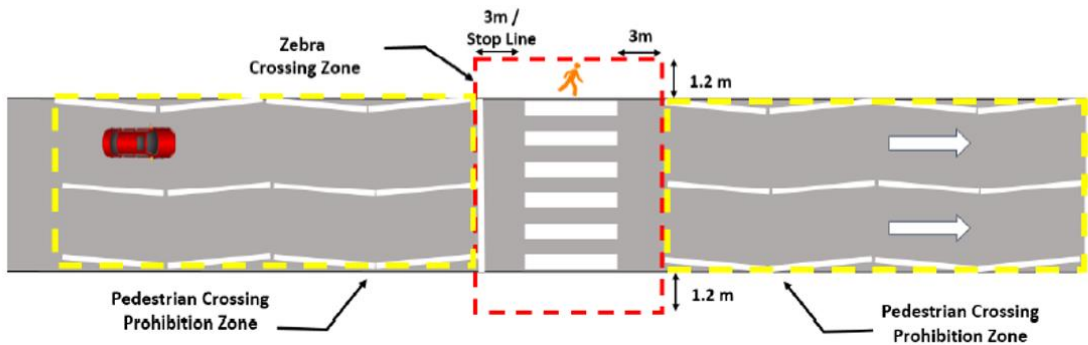

Figure 1: Zebra crossing zone defined in TR68-1 [1]

### 3.2.2.5 Yellow Box

Table 23: Consolidated guidelines from TR68-1, BTD & FTD for AV behaviour at yellow-box

| Yellow-box | Parameters |
|---|---|
| **BTD 93**: It is an offence for any driver to drive his/her vehicle into a junction marked with a yellow box and cause obstruction even if the lights are in his/her favour. However, there are certain situations in which a driver or rider will not be penalised for entering the yellow-box junction | |

### 3.2.2.6 Bus-stop

There is currently no defined expected behaviour for AVs at bus-stops in current TR68-1, BTD and FTD, only regarding lane markings and behaviour near a bus-stop as shown in Table 24 below. This has been identified as missing guidelines for buses. Currently, developers interested in this field have to consult with bus operators on expected behaviours.

Table 24: Consolidated guidelines from TR68-1, BTD & FTD for AV behaviour at bus-stops.

| At Bus-stops | Parameters |
|---|---|
| **BTD 53:** Vehicles other than omnibuses, bicycles, emergency service and police vehicles must avoid using the bus lanes during restricted hours. Non-scheduled buses such as school and factory buses may use the bus lane. However, they are not allowed to stop, pick up or let their passengers alight along bus lanes.<br>(a) **Normal Bus Lane**: A continuous yellow line and a short horizontal line at intervals indicate a normal bus lane. No driving or parking of vehicles in the bus lane during the bus lane operational hours – Mondays to Fridays: 7.30am to 9.30am and 5.00pm to 8.00pm, except on Saturdays, Sundays & Public Holidays. Vehicles can use the bus lane where there is a broken yellow line to turn into a side road or turn from a side road into the bus lane road.<br>(b) **Full-Day Bus Lane:** An additional red line marked parallel to the yellow line indicates a full-day bus lane. No driving or parking of vehicles in bus lane during the full-day bus lane operational hours – Mondays to Saturdays: 7.30am to 11.00pm, except on Sundays & Public Holidays.<br>c) **Bus Priority Box** (also known as Mandatory Give-Way to Buses): The section in yellow indicates a mandatory give-way to buses area. Slow down and watch out for buses pulling out of the bus bay. Stop before the give way line and give way to buses exiting the bus bay at the location.<br>(d) **Enhanced Chevron Zone at Bus-Stop Separator:** Bus-stop separator areas are painted red to alert motorists of the presence of merging vehicles from bus-stops. No driving or parking of vehicles in the enhanced chevron zone. | |
| **FTD 223:** At a bus stop, be alert and prepared to stop for pedestrians, especially school children, as they may dash out from the front of a bus or between buses. | |

### 3.2.3  Driving actions/ Manoeuvre related actions

#### 3.2.3.1    *Driving in Lane with Traffic*

##### 3.2.3.1.1.    <u>Following Actor</u>

Table 25: Consolidated guidelines from TR68-1, BTD & FTD for driving in lane with traffic – following other vehicles.

| Driving in lane with traffic (Following) | Parameters |
|---|---|
| **TR68-1:** 6.3.4) The AV shall maintain a safe car following distance (BTD 135) to all moving vehicles in the same lane "ahead" with "established lane presence". A vehicle will be considered as "ahead" in lane if their front bumper is further along the lane in a longitudinal sense than the AV's front bumper. A vehicle will be considered to have "established lane presence" if the majority of their footprint is within the lane boundaries. The safe following distance shall be determined so that if the leading vehicle would initiate a complete stop, the AV would be able to stop behind that vehicle with a longitudinal distance of at least 2 m. The AV should respond proactively when other vehicles are observed to be cutting into its lane so as to prevent or minimise violation the safe following distance as much as possible. | |
| **BTD 135**: To be able to stop with an appropriate space between your vehicle and the vehicle in front, you must allow at least one car length<br>for every 16km/h of your speed. | |
| **BTD 136:** A simple way to ensure a safe gap between you and the vehicle in front is to use the 'two-second' rule. As the vehicle in front of you passes a stationary object, such as a lamp post on the side of the road, start counting to yourself 'one-thousand-and-one, one thousand-and-two'. This will take you 2 seconds. If you reach the same spot before you finish these eight words, you are following too closely and it is necessary to slow down. This rule will keep you at a safe following distance and will apply to all types of vehicles at any speed. | |

### 3.2.3.1.2. *Stationary Actor*

Table 26: Consolidated guidelines from TR68-1, BTD & FTD for driving in lane with traffic – with stationary actors.

| Driving in lane with traffic (Following) | Parameters |
|---|---|
| **TR68-1:** 6.3.2.c) Whenever the AV is in motion with speed less than 30 km/h, the lateral clearance to pedestrians on the road surface and not laterally moving away from the AV's path shall be no less than 1 m. | Lateral Clearance: > **1.0m** |
| **TR68-1:** 6.3.2.d) For AV motion at speeds equal to or above 30 km/h, the lateral clearance to pedestrians on the road surface and not laterally moving away from the AV's path shall be no less than 1.5 m. | Lateral Clearance: > 1.5m |
| **FTD226**: When passing fixed obstacles, keep a gap of at least 0.5 m from them | 0.5m |

### 3.2.3.1.3. Jaywalking

Table 27: Consolidated guidelines from TR68-1, BTD & FTD for driving in lane with traffic – with stationary actors.

| Driving in lane with traffic (Jaywalking) | Parameters |
|---|---|
| **TR68-1:** 6.3.2.a) The AV shall not collide with any pedestrian, cyclist, or PMD user Threshold: None | |
| **TR68-1:** 6.3.2.b) The AV shall maintain sufficient clearance to all pedestrians, where sufficient clearance is dependent also on the AV speed. | None |
| **TR68-1:** 6.3.2.e) For any AV speed, the lateral clearance for pedestrian on the road surface and laterally moving away from the AV's path, shall be no less than 1 m. | Lateral Clearance: > 1.0m |
| **TR68-1:** 6.3.2.f) For any AV speed, the lateral clearance for pedestrians not on the road surface and laterally moving away from the AV's path, shall be no less than 0.5 m. | Lateral Clearance: > 0.5m |
| **BTD 148:** Pedestrians using rural roads tend to be less "road-wise". Watch out for them | |
| **BTD 149:** When you are passing a pedestrian or cyclist, always maintain as wide a distance from him/her as possible and drive slowly. | |
| **BTD 150:** When driving across a muddy section of the road or through a puddle, you should slow down to avoid splashing muddy water onto pedestrians. | |
| **BTD 167:**<br>(a) Always be patient. Do not rush or lose your cool on the road.<br>(b) It is not gracious to show any provocative gesture when other drivers do something wrong or cause inconvenience to you. No one would deliberately get himself/herself involved in an accident. | |
| **FTD 223:** At a bus stop, be alert and prepared to stop for pedestrians, especially school children, as they may dash out from the front of a bus or between buses. | |
| **FTD 228:** Accidents involving pedestrians often result in serious injury or even death. Pedestrians, especially the very young and the very old, are the most vulnerable group of road users. It is your duty as a driver to protect pedestrians. | |
| **FTD 229:** Young children below the age of fifteen are prone to accidents. This is because they are impulsive, playful, curious and unaware of the dangers on the road. Be very careful when driving near schools, playgrounds and in residential areas. Be especially alert when driving near ice-cream sellers and roadside vendors. Always keep a look-out for children who may run or dash across the road. | |
| **FTD 230:** The old and the handicapped are vulnerable to accidents on the roads. It is a known medical fact that upon reaching the age of 50 years, a person's<br>(a) vision begins to fade;<br>(b) hearing is impaired;<br>I body weakens and reflexes become slower.<br>Aged and handicapped persons are usually slow in reacting to traffic situations. Therefore when approaching the old or the handicapped, give them plenty of room and stop if necessary for them to cross the road. | |
| **FTD 231:** Jaywalkers cross the road anywhere they please, heedless of traffic rules and vehicles. Be on the look-out for them. | |
| **FTD 232:** Joggers may run or dash across the road unexpectedly without any regard for vehicles on the road. Always be alert and give way to them if necessary. | |
| **FTD 239:** When you are turning into a side road, look out for pedestrians. | |
| **FTD 240:** Motorists turning left to join the main traffic from a side road should look out for pedestrians from the left. | |
| **FTD 246: When turning**<br>(a) check your mirrors and blind spots;<br>(b) look out for cyclists between you and the kerb;<br>do not make a sharp turn as you may knock the cyclist down. Slow down and give way to the cyclist if it is not safe to turn. | |

### 3.2.3.2 Ego Lane Change

*A driving manoeuvre where the Ego Vehicle would need to change lane to the right or left of its current lane.*

Table 28: Consolidated guidelines from TR68-1, BTD & FTD for an AV carrying out lane change.

| Lane Change | Parameters |
|---|---|
| **TR68-1: 7.5** If the AV plans a lane change in the absence of an immediate safety concern, the lane change manoeuvre shall commence not before 3seconds have passed after the activation of the signal indicator to the intended side of lane change as defined in 5.6.4.6.4 of ECE/TRANS/WP.29/GRRF/85 and BTD 131. The lane change manoeuvre stars when the AV stars to move laterally so that it gets closer to the lane to which it intended to change. The lateral movement to approach the lane marking and the lateral movement necessary to complete the lane change manoeuvre should be completed as one continuous movement. **BTD131:** Always give clear signals well in advance of your intentions (at least 3 seconds) before your manoeuvre so that other road users can interact safely | 3 seconds |
| **TR68-1: 7.5a)** Before commencing a lane change manoeuvre, the AV shall ensure that the manoeuvre does not force any other road user (ORU) to take potentially unsafe evasive action. | |
| **TR68-1: 7.5b)** The AV shall keep a safe following distance at all times before, during and at the end of any lane change manoeuvre. | |
| **TR68-1: 7.5c)** The AV shall cancel its signal after the lane change manoeuvre is completed, i.e. its footprint is enclosed by the boundaries of the new lane. | |
| **FTD 106:** The most difficult aspect of changing lanes is the estimation of speed and distance of vehicles approaching from behind. As such, drivers will have to confirm the situation with regular glances at the mirror inside the cabin as well as the wing mirror. To change lanes safely, this is what you should do:<br>At 'a'<br>(i) Check mirrors;<br>(ii) Signal your intention;<br>(iii) Check blind spot<br>At 'b'<br>(i) Adjust your speed;<br>(ii) You may have to slow down or speed up depending on the traffic situation behind you.<br>At 'c'<br>When it is safe, accelerate smoothly and steer gently into the lane intended without interrupting the flow to traffic.<br>At 'd'<br>Cancel your signal and resume your normal speed. | |
| **FTD107:** You should not change lanes or cross the centre dividing line unnecessarily | |
| **FTD108**: Always keep well to the left when driving along two-way streets or dual carriageways, unless your path of travel is obstructed by road works, parked vehicles, etc. In such circumstance, you may cross the centre line or move to the lane on your right. On doing so, take care to ensure that your intended path is safe and clear before you move to the right. | |
| **FTD109**: Where there are two lanes, the left lane is for normal driving and the right lane is for overtaking and turning right. | |
| **FTD110**: Where there are three lanes, the left lane is for slower vehicles, the centre is for faster vehicles and right lane is for vehicles overtaking and turning right. | |

### 3.2.3.3  Overtaking

#### 3.2.3.3.1.   Overtaking- ALL (General)

Table 29: Consolidated guidelines from TR68-1, BTD & FTD for driving in lane with traffic – overtaking.

| Overtaking -ALL (General) | Parameters |
|---|---|
| **TR68- 7.6 a)** An AV shall only perform an overtaking manoeuvre when it is safe to do so. | |
| **TR68-1 7.6 b)** An AV shall check for any traffic signs prohibiting overtaking before initiating an overtaking manoeuvre | |
| **TR68-1 7.6 b)** The AV shall not persist in overtaking if the ORU in front of increases its speed | |
| **TR68-1 7.6 e)** An AV shall adopt the following steps when performing an overtaking manoeuvre:<br>Signal right, Right lane change, Accelerate, Signal left, Left lane change, Cancel signal, Resume normal speed | |
| **TR68-1 7.6 f) i)** An AV shall always overtake to the right, except when the vehicle in front has signalled his/her intention to turn right | |
| **TR68-1 7.6 g) iv) v)vi)vii)**   An AV shall not perform an overtaking manoeuvre at, or when approaching the following:<br>      iv) The brow of a hill<br>      v) Double white lines<br>      vi) On a narrow bridge<br>vii) On a steep slope or incline | |
| **TR68-1 7.6 j** When being overtaken, an AV shall slow down to allow the overtaking ORU to pass if so, required for safety reasons. | |
| **BTD 71** After overtaking, return to the appropriate lane on the road as soon as it is safe to do so, but do not cut in sharply in front of the vehicle you have just overtaken | |
| **BTD 137 (f)** Do not overtake more than one vehicle at a time; | |
| **BTD 137 c)** Do not try to overtake others when you are being overtaken; | |
| **BTD150** When you are passing a pedestrian or cyclist, always maintain as wide a distance from him/her as possible and drive slowly. | |
| **FTD226** When passing fixed obstacles, keep a gap of at least 0.5 m from them | 0.5m to fixed obstacles |
| **FTD227** When passing moving vehicles, keep a gap of at least 1.5 m from them | 1.5m to moving vehicles |

#### 3.2.3.3.2.   Overtaking -Stationary

Table 30: Consolidated guidelines from TR68-1, BTD & FTD for driving in lane with traffic – overtaking stationary actors.

| Overtaking / Stationary / single lane in ego direction | Parameter |
|---|---|
| **TR68-1 6.3.2.c)** Whenever the AV is in motion with speed less than 30 km/h, the lateral clearance to pedestrians on the road surface and **not laterally moving away** from the AV's path shall be no less than 1 m. | Threshold: 1m |
| **TR68-1 6.3.2.d)** For AV motion at speeds equal to or above 30 km/h, the lateral clearance to pedestrians on the road surface and **not laterally moving away** from the AV's path shall be no less than 1.5 m. | Threshold: 1.5m |
| **TR68-1 7.6 d)** An AV shall only move into the lane for oncoming traffic when performing an overtaking manoeuvre if the intended path is safe and clear of traffic | |
| **TR68-1 7.6 c) i)** When performing an overtaking manoeuvre, an AV shall never cross the doble while line , except when Its path of travel is obstructed by road works | |
| **TR68-1 7.6 c) ii)** When performing an overtaking manoeuvre, an AV shall never cross the doble while line , except when Its path of travel is obstructed by parked vehicles. | |
| **FTD 136** When overtaking you have to, estimate not only the space ahead of you but also the speed and distance of oncoming vehicles | |
| **FTD 221:** When passing by a parked vehicle, look out for drivers or passengers opening the doors of their vehicles and keep a safe gap of about 1 meter between you and the parked vehicles. | Threshold: 1.0m |

### 3.2.3.3.3. *Dynamic – Single lane*

Table 31: Consolidated guidelines from TR68-1, BTD & FTD for driving in lane with traffic – with dynamic actors.

| Overtaking / Dynamic / single lane in ego direction | Parameters |
|---|---|
| **TR68-1 7.6 d)** An AV shall only move into the lane for oncoming traffic when performing an overtaking manoeuvre if the intended path is safe and clear of traffic | |
| **TR68-1 7.6 i) & 6.3.3)** Before during and after overtaking a cyclist, an AV shall keep at least 1.5 m lateral distance from the cyclist. | Threshold: 1.5m |
| **TR68-1 7.6 h (i)** An AV shall not perform an overtaking manoeuvre when the ORU in front is about to overtake another vehicle in front of it | |

### 3.2.3.3.4. *Dynamic – Multi-lane*

Table 32: Consolidated guidelines from TR68-1, BTD & FTD for driving in lane with traffic – overtaking dynamic actors.

| Overtaking / Dynamic / multi-lane in ego direction | Parameters |
|---|---|
| **TR68-1 7.6 f) iii)** An AV shall always overtake to the right, except in these scenarios when the ORUs on the right lane are moving slower than an AV is | |
| **TR68-1 7.6 f) iv)** An AV shall always overtake to the right, except on one-way streets (but not dual carriageways) where vehicles may pass on either side | |
| **TR68-1 7.6 h (i)** An AV shall not perform an overtaking manoeuvre when the ORU in front is about to overtake another vehicle in front of it | |
| **TR68-1 7.6 h) ii)** An AV shall not perform an overtaking manoeuvre when the ORU in front is changing from the left to the right lane | |

### 3.2.3.4 Occlusion

Table 33: Consolidated guidelines from TR68-1, BTD & FTD for occlusion situations

| Resolving occlusion | Parameters |
|---|---|
| **TR68- 6.1.c:** Speed should account for the road surface conditions and current and upcoming range of object sensor detection and visibility to other road users. When faced with areas of occlusions or range restrictions, an AV shall react to reasonable worst-case assumptions of actors being present just outside of field of view. In the event that occlusions will not be satisfactorily removed without forward motion, the AV should be allowed to creep forward at a reduced speed until the occlusions have been removed | |
| **BTD 148:** Watch for pedestrians who come out suddenly from behind stationary vehicles and other obstructions. Be very careful near schools and bus stops. | |

# 4. Analysis of Identified Situations

Some traffic situations were identified for further analysis to focus on relevant guidelines in specific situations that could not be generally applied. This also allows to detailed assessment if generic guidelines are relevant for such situations. These situations were identified from CETRAN's experience from assessments and previous collaborations that highlighted challenges with either missing guidance or difficulty for AV developers to apply these guidelines. In this following section, the overview of the analysis method will be described followed by detailed analysis of a few selected situations. The overview of the analysis method is detailed in 4.1. From Section 4.2 to 4.4, example situations analysed by CETRAN team is detailed. Examples from section 4.5.1 to 4.5.3 are from the collaboration with STEAS. All the recommended additional guidelines from these examples are applicable for both Class 3 and Class 4. The additional recommended guidelines and traffic situations analysed form the second part of the behaviour database for AVs. Some parameters were tested on the CETRAN track to check on the practicality of the recommendations.

## 4.1 Overview of analysis method

The overview of the situation analysis follows five points as listed below. It is broken into two major sections, firstly on the general situation and second in the corresponding analysis. The first step would be to briefly describe the identified traffic situation and what is generally happening here. The assumption within this situation should also be listed to focus the analysis within some limitations. Missing guidance of the situation can be identified in the third step, after checking current guidelines and from CETRAN and STEAS experience. To begin with the analysis, one step CETRAN has incorporated is listing the decision steps that an AV needs to consider when implementing the actions within the situation. This leads to the fifth step of identifying the current guidelines at each decision step and corresponding missing and recommended guidelines.

1. General Situation – Description
2. General Situation – Assumptions
3. General Situation – Missing guidance
4. Analysis of Situation – Decision steps
5. Analysis of Situation– Expected behaviour (Current guidelines with additional suggestions)

## 4.2  Example 1- Left turn slip road

This first example of an AV or ego vehicle approaching a left turn slip road at a traffic junction was used to facilitate discussions at an workshop with AV developers.

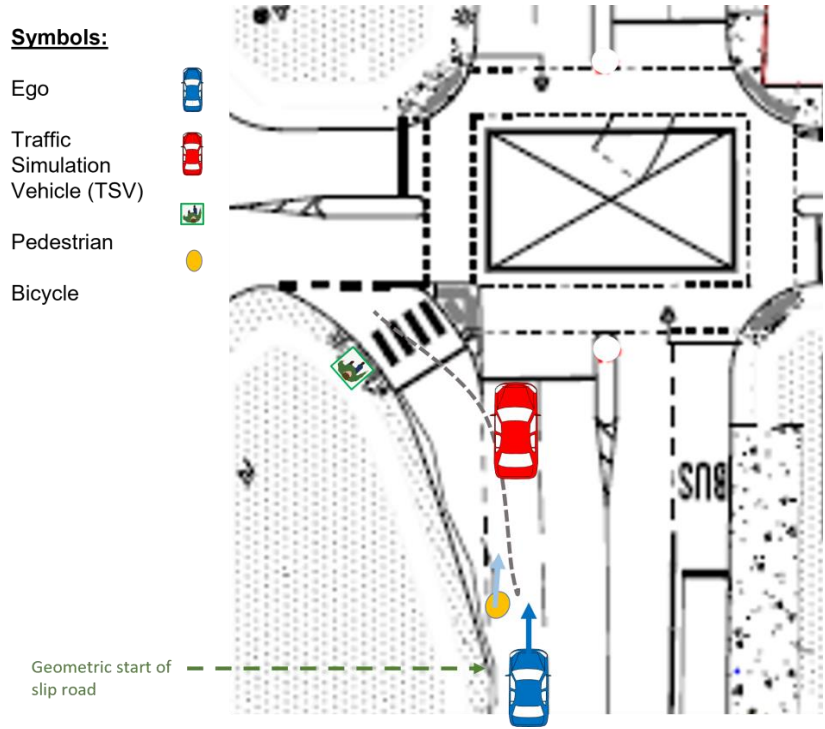

Figure 2: Traffic situation where an AV (ego vehicle) is approaching a left turn slip road with other traffic actors depicted.

### 4.2.1  General Situation

**Description:**

- Ego vehicle to turn left at left turn slip road, with moving cyclist ahead, pedestrian approaching and crossing zebra crossing, occlusion of view of approaching traffic from the right (static Traffic Simulated Vehicle (TSV)).

**Assumptions:**

- Road layout as per diagram – not a school zone, no police hand signalling

**Missing guidance**

- When to overtake cyclist / stay behind?
- Trajectory around cyclist
- Trajectory into left turn slip road
- When to stop / when not to stop at zebra crossing
- Resolving occluded view at left turn slip road

### 4.2.2 Analysis of situation

#### *4.2.2.1 Decision Steps Required*

1. If staying behind cyclist- distance to stay behind

2. **Decision of overtaking cyclist or not**
    - Traffic in adjacent lane[1]?
    - Clear space ahead of cyclist? (no traffic ahead)
    - Upcoming manoeuvre (imminent left turn)
    - Not to inconvenience cyclist

2. **If overtake cyclist –lane change and overtaking procedure:**
    - Signal indicator to show intent to overtake.
    - Trajectory around cyclist (both before and after cyclist), straddle lane or fully move into adjacent lane
    - Clearance to cyclist
    - Speed of manoeuvre

4. **Entering left turn slip road**

    - **Signal intention to turn left at slip road**

    - **Location along slip road to cross into lane (showing clear intention of manoeuvre to other road users)**

5. **Stopping at zebra crossing**
    - When to stop (pedestrian location and direction of travel)
    - Where to stop relative to stop line
    - When to move off (pedestrian didn't cross, pedestrian location along crossing)
6. **Stopping at give way line**
    - Not to obstruct zebra crossing (long vehicle, short distance to give way line)
    - Need to stop or not

7. **Resolving occluded view**

#### *4.2.2.2 Guidelines of each decision step*

##### *4.2.2.2.1. Following Cyclist*

Within this one action, there are multiple guidelines seen in Table 34 below. 2 seconds following from BTD 136 [2] incorporates element for human reaction time during drive which may not be relevant for AVs and a system latency time could be used instead. This value is also more conservative than 1 car length per 16km/h of vehicle speed (BTD 135) [2].

---

[1] Adjacent lane means the lane that is nearest to the AV, having a common boundary (lane marking) with the lane which the AV is in.

Table 34: Current guidelines from TR68-1, BTD and FTD where applicable on following distances

| Driving in lane with traffic (Following) | Parameters |
|---|---|
| **TR68-1:** 6.3.4) The AV shall maintain a safe car following distance (BTD 135) to all moving vehicles in the same lane "ahead" with "established lane presence". A vehicle will be considered as "ahead" in lane if their front bumper is further along the lane in a longitudinal sense than the AV's front bumper. A vehicle will be considered to have "established lane presence" if the majority of their footprint is within the lane boundaries.<br>The safe following distance shall be determined so that if the leading vehicle would initiate a complete stop, the AV would be able to stop behind that vehicle with a longitudinal distance of at least 2 m. The AV should respond proactively when other vehicles are observed to be cutting into its lane so as to prevent or minimise violation the safe following distance as much as possible. | Can stop 2m behind |
| **BTD 135:** To be able to stop with an appropriate space between your vehicle and the vehicle in front, you must allow at least one car length for every 16km/h of your speed. | 1 car length for every 16 km/h |
| **BTD 136:** A simple way to ensure a safe gap between you and the vehicle in front is to use the 'two-second' rule. As the vehicle in front of you passes a stationary object, such as a lamp post on the side of the road, start counting to yourself 'one-thousand-and-one, one thousand-and-two'. This will take you 2 seconds. If you reach the same spot before you finish these eight words, you are following too closely, and it is necessary to slow down. This rule will keep you at a safe following distance and will apply to all types of vehicles at any speed. | 2seconds |
| **TR68-1:** 6.3.2.a) The AV shall not collide with any pedestrian, cyclist, or PMD user | |

During the workshop, the question was posed to AV developers that with multiple guidelines above seen in Table 34 above, if they took the most conservative guideline or does adoption depend on situation. One response noted was that following the most conservative guideline of 2 seconds following could leave a gap that encourages cutting-in instead of the desired safety principle. It was also noted that TR 68-1 6.3.4 was established to allow for AV vehicle capability and simply putting a 2metre distance when stopped allows for developers to calculate desirable distances based on each vehicle's capability. CETRAN has also referred to metrics from International Organization for Standardization 15622:2018 (ISO) such as adaptive cruise control requirements [6]. The minimum time gap in ISO 15622:2018 was 0.8seconds to be slightly smaller distance than a 1 car length per 16km/h guideline from BTD using a standard car length.

The various safe distances from the guidelines (TR68-1, BTD 135 and BTD136) could not be applied concurrently as they are conflicting. This was tested at CETRAN test track to visualize the various distances in the guidelines, and is seen in Figure 3 below. CETRAN recommends the concept from TR68-1 "so that if the leading vehicle would initiate a complete stop, the AV would be able to stop behind that vehicle with a longitudinal distance of at least 2 m " [1] to be adopted for AV assessment on safe following distance, with the consideration that developers should incorporate system reaction time or latency and desirable gentle braking rates. It was deemed to be a safe distance by former TP member and considering AV developers feedback from the workshop.

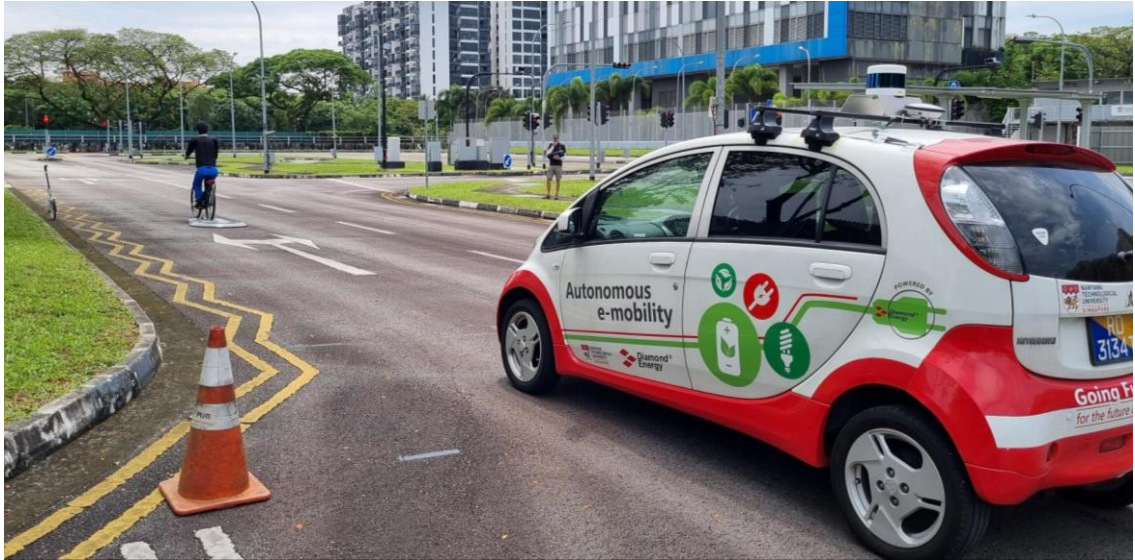

Figure 3: Tests on CETRAN track regarding following distance behind other road users (ORU)

### 4.2.2.2.2. *Decision of overtaking cyclist*

In the first step in deciding on overtaking cyclist ahead before a left-turn slip road, some considerations are to check for traffic in adjacent lane, sufficient distance ahead of the cyclist, and imminent manoeuvre of entering of left turn slip road after overtaking such that it does not inconvenience the cyclist. The relevant guidelines for this decision have been selected in Table 35 below.

Table 35: Current guidelines from TR68-1, BTD and FTD where applicable on overtaking

| Overtaking decision | Parameters |
|---|---|
| **TR68-1:** 7.6 g)ii) An AV shall not perform an overtaking manoeuvre at, or when approaching a road junction | *How much before?* |
| **TR68-1:** 7.10.2 c) An AV shall signal its intention at least 3 seconds before entering the lane which permits the intended direction of turn at a junction or a slip road. | 3 sec |
| **TR68-1: 7.5** If the AV plans a lane change in the absence of an immediate safety concern, the lane change manoeuvre shall commence not before 3seconds have passed after the activation of the signal indicator to the intended side of lane change as defined in 5.6.4.6.4 of ECE/TRANS/WP.29/GRRF/85 and **BTD 131**. The lane change manoeuvre starts when the AV stars to move laterally so that it gets closer to the lane to which it intended to change. The lateral movement to approach the lane marking and the lateral movement necessary to complete the lane change manoeuvre should be completed as one continuous movement.<br>**BTD131:** Always give clear signals well in advance of your intentions (at least 3 seconds) before your manoeuvre so that other road users can interact safely | 3 sec |

For lane change, it was noted that developers signal for 3 seconds before entering the left turn slip road as desired in TR68-1: 7.5 (BTD 131), using the guidance that lane change manoeuvre should commence after 3 seconds of activating the signal. However, when there is an overtaking manoeuvre just before entering the left turn slip road, this is not defined from TR68-1: 7.6 g)ii when this overtaking manoeuvre should be completed. CETRAN believes a metric either in distance or time from the road junction should complement TR68-1: 7.6 g)ii. CETRAN recommends that the overtaking manoeuvre be completed at least 3 seconds before reaching a traffic junction (the left turn slip road in this situation). The recommendation complies with TR68-1 7.5 [1] and give additional metrics to TR68-1: 7.6 g)ii on how much before a traffic junction the overtaking manoeuvre should be completed.

### 4.2.2.2.3. *Overtaking cyclist: Lane Change and Overtaking Procedure*

When an AV has decided on overtaking the cyclist in a similar traffic situation, it should follow the guidelines that are found in Table 36 below when carrying out a lane change action, and Table 37 for overtaking procedure as a lane change is required when overtaking.

Table 36: Current guidelines from TR68-1, BTD and FTD where applicable on lane change AND interaction with other actors.

| Lane Change | Parameters |
|---|---|
| **TR68-1: 7.5** If the AV plans a lane change in the absence of an immediate safety concern, the lane change manoeuvre shall commence not before 3seconds have passed after the activation of the signal indicator to the intended side of lane change as defined in 5.6.4.6.4 of ECE/TRANS/WP.29/GRRF/85 and **BTD 131**. The lane change manoeuvre stars when the AV stars to move laterally so that it gets closer to the lane to which it intended to change. The lateral movement to approach the lane marking and the lateral movement necessary to complete the lane change manoeuvre should be completed as one continuous movement. **BTD131:** Always give clear signals well in advance of your intentions (at least 3 seconds) before your manoeuvre so that other road users can interact safely | 3 sec |
| **TR68-1: 7.5a)** Before commencing a lane change manoeuvre, the AV shall ensure that the manoeuvre does not force any ORU to take potentially unsafe evasive action. | *What is unsafe?* |
| **TR68-1: 7.5b)** The AV shall keep a safe following distance at all times before, during and at the end of any lane change manoeuvre. | *What distance is safe?* |
| **TR68-1: 7.5c)** The AV shall cancel its signal after the lane change manoeuvre is completed, i.e. its footprint is enclosed by the boundaries of the new lane. | |
| **BTD 150** When you are passing a pedestrian or cyclist, always maintain as wide a distance from him/her as possible and drive slowly. | *What is as wide as possible?* TR68-1 7.6 i) & 6.3.3) |

Table 37: Current guidelines from TR68-1, BTD and FTD where applicable on overtaking

| Overtaking (applicable to ALL generic) | Parameters |
|---|---|
| **TR68-1 7.6 a)** An AV shall only perform an overtaking manoeuvre when it is safe to do so. | |
| **TR68-1 7.6 b)** An AV shall check for any traffic signs prohibiting overtaking before initiating an overtaking manoeuvre | |
| **TR68-1 7.6 d)** The AV shall not persist in overtaking if the ORU in front of it increases its speed | *Does this apply to a cyclist?* |
| **TR68-1 7.6 e)** An AV shall adopt the following steps when performing an overtaking manoeuvre: Signal right, Right lane change, Accelerate, Signal left, Left lane change, Cancel signal, Resume normal speed | |
| **TR68-1 7.6 i) & 6.3.3)** Before during and after overtaking a cyclist, an AV shall keep at least 1.5 m lateral distance from the cyclist. | 1.5m |
| **BTD 71** After overtaking, return to the appropriate lane on the road as soon as it is safe to do so, but do not cut in sharply in front of the vehicle you have just overtaken | *Definition of cut in sharply?* |

Developers were asked at the workshop if reference such as The United Nation Economics Commission for Europe (UNECE) 79 for lateral acceleration was used in development when defining what is a sharp lane change where lateral acceleration of 3m/s$^2$ for some vehicle types, and jerk values below 5m/s$^3$ over an average of 0.5seconds [7]. Some developers indicated that they did use as limits in designing for AV behaviour, especially on passenger carrying vehicles where ride comfort is pertinent.

Overall, the guidelines are clear on what is required for overtaking procedure in TR68-1 7.6 e) and required lateral clearances to maintain. However, there are no parameters in TR68-1 7.5a) and b) on what constitutes as 'safe distance' that will not force an other road user (ORU) to take potentially unsafe evasive action and which is a safe following distance. CETRAN has referred to metrics from ISO 17387 on lane change decision aid systems [7]. On lane change recommendations, CETRAN considered possibilities of the TSV/cyclist approaching from behind in the adjacent lane to be both faster and slower than the AV. In the case where the AV and or TSV is travelling at a high speed, it is preferable to use a more conservative value, and the maximum between 1 car length per 16km/h of the TSV/cyclist speed or 9 seconds TTC allows for this safe conservative approach. This safe distance is the longitudinal distance between the back of the AV to the front of the TSV/cyclist approaching from behind in the adjacent lane.

In view of potential ambiguity for the development and assessment of AVs, CETRAN has suggested the following guideline:

1. When changing lanes, the AV should be maintaining a safe distance of 1 car length per 16km/h of the TSV/cyclist speed or 9[2] seconds TTC whichever is bigger to the TSV/cyclist coming from behind in the adjacent lane (i.e. when the AV starts to cross the lane marking). (Refer to Figure 4 left below).
2. If there is an oncoming vehicle (vehicle ahead travelling towards ego) in the adjacent lane during the overtaking manoeuvre, the AV should change back into its original lane with at least a 2 second time gap[2] (time to collision (TTC)) with the TSV/cyclist oncoming in the adjacent lane. (Refer to Figure 4 right below)
3. A recommended value for lateral acceleration should be used in TR68-1, such as UNECE 79 5.6.2.1.3 [8] according to vehicle class and vehicle speeds, such that bad behaviour of 'sudden cut ins' can be better defined, together with safe distances to other actors. This is recommended to follow a maximum of 3m/s$^2$ for speeds between 10-60km/h (M1, N1 vehicle category) and with a moving average over half a second of the lateral jerk generated to not exceed 5m/s$^3$ [8].

This would allow for a clearer definition for development of acceptable AV behaviour and the corresponding assessment for AVs in Singapore for lane change action applicable within an overtaking procedure.

---

[2] This was tested on CETRAN track within track limits.

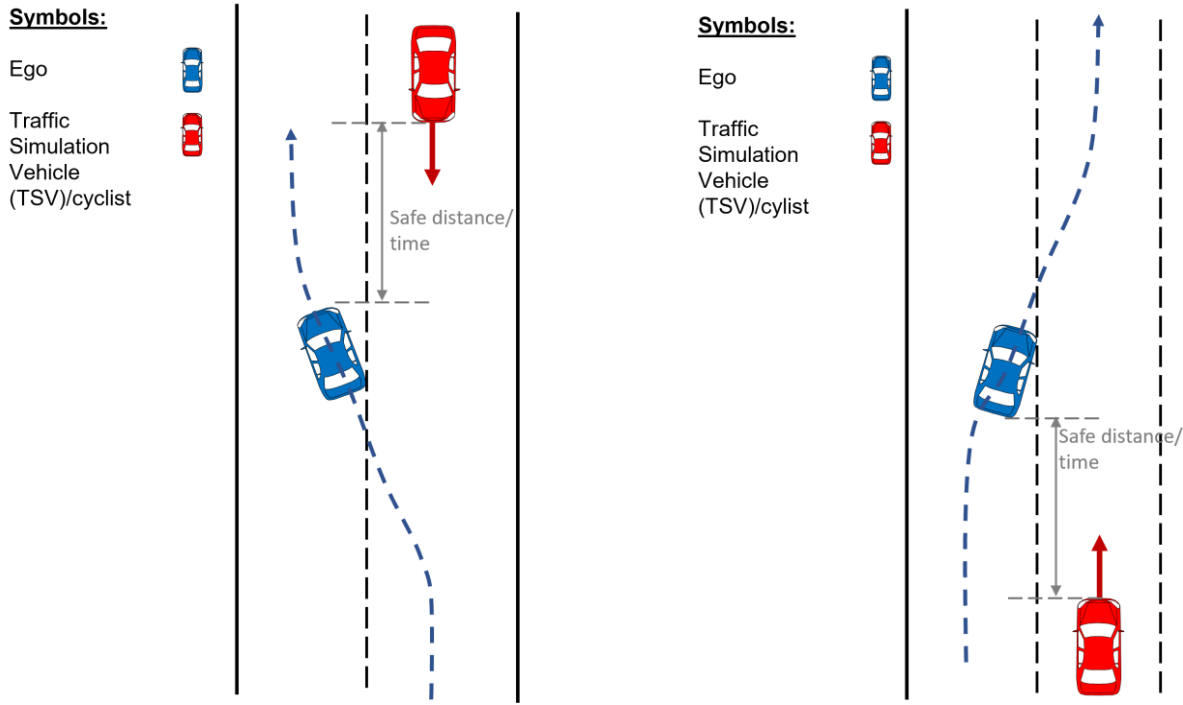

Figure 4: Sketch on **left:** to illustrate safe distance/ time when Ego is changing lanes with TSV/cyclist approaching from behind in adjacent lane. On **Right:** Sketch of safe distance/time between Ego and TSV/cyclist when Ego is changing lane, to oncoming TSV/cyclist in adjacent lane.

### 4.2.2.2.4.    *Entering Left turn slip road*

Table 38: Current guidelines from TR68-1, BTD and FTD where applicable on entering left turn slip road

| TURNS-Left Turn Slip Road | |
|---|---|
| **TR68-1: 7.10.2 a)** When turning left or right at a junction, signalised or non-signalised, an AV shall keep to its lane corresponding to its lane before turning. | |
| **TR68-1: 7.10.2 b)** An AV shall enter the lane that permits the intended direction of turn at a junction or a slip road as indicated by road markings. | |
| **TR68-1: 7.10.2 c)** An AV shall signal its intention at least 3 seconds before entering the lane which permits the intended direction of turn at a junction or a slip road. | 3 seconds |

When entering a left turn slip road, the current guidance available is regarding signalling due to lane change or if lane is appropriate to enter as seen in Table 38 above. However, there have been instances in CETRAN's experience where 'late entrance[3] into left turn slip road' has been observed. CETRAN has then noted that additional guidance is required to align assessors and developers on acceptable behaviour into left turn slip road, when is determined as acceptable behaviour such as timing to enter slip road or perhaps by determining appropriate distance to maintain to kerb. The idea of maintaining a distance to the kerb would not apply when there are step changes in lane as observed on some public left turn slip roads. This was trialled on CETRAN test track to visualize the recommended additional guideline.

---

[3] "late entrance" refers to the undesirable behaviour where the AV cuts the left turn slip road near the zebra crossing

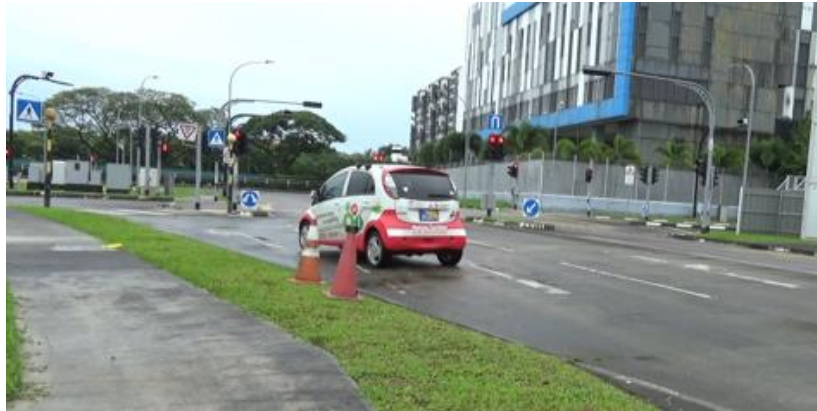

Figure 5: Test at CETRAN track for left turn slip road recommended location to start entering lane.

CETRAN's recommendation for additional guidance would be the following:

1. When entering a left turn slip road with a zebra crossing approaching a traffic junction: An AV should start turning at 3metre after the start of the geometric start of slip road (see Figure 2 above), maintaining a smooth trajectory from the centre of the original lane into the centre of the left turn slip road lane, so as not to encourage motorcyclists from entering from the left of the AV. (This is applicable for urban roads and not for exits from highways, and not for special cases where the entrance is very short).

### *4.2.2.2.5.    Stopping for Zebra Crossing*

Table 39: Current guidelines from TR68-1, BTD and FTD where applicable on zebra crossing

| Overtaking | Parameters |
|---|---|
| **TR68-1 7.9.1** Giving right of way.<br>An AV shall give right of way to VRU at pedestrian crossings, including zebra crossing unless they have no intent to cross [ BTT 143] [FTT233] | *See below TR68 – 7.9.6 (5 sec)* |
| **TR68-1 7.9.5** Assumption of pedestrians' intent occupying zebra crossing zone [BTT 145 a] [FTT236 A]<br>VRU who are occupying the zebra crossing zone and are not moving away from the path of the AV shall also be assumed to have intent to cross at least until the AV comes to a complete stop at the pedestrian crossing | *How long to wait? See below TR 68-1 7.9.6 (5 sec)*<br><br>*Pedestrian Zone – in TR68-1 (see Figure 6 below)* |
| **TR68-1 7.9.2** Approaching a pedestrian crossing<br>When approaching a pedestrian crossing, which includes zebra crossing, the AV should adjust its speed such that it can reasonably stop **[ BTT 51d/e]** | |
| **TR 68-1 7.9.6** AV's speed limit when traversing the zebra crossing zone. The AV shall traverse the zebra crossing zone at the speed of not exceeding 30km/h when the zebra crossing zone is occupied by a VRU who has not displayed intent to cross within the passing of 5 seconds | 30 km/h;<br>5 sec |
| **TR68-1 7.9.4** Stopping at Zebra crossing<br>AV shall stop outside the zebra crossing zone when VRU are present inside the zebra crossing zone with one or more of the following attributes:<br>    a) On the road surface less than 4 m away laterally from the AV footprint<br>    b) Moving towards the AV's path and is within 7m laterally from the AV footprint<br>    c) Has displayed the intent to cross the AV's path and is within 7m laterally from the AV 's footprint<br>AV shall stop outside the zebra crossing zone at the stop line in accordance with TR68-1 6.4 or stopping at stop lines.<br>When stopping at stop lines the AV should achieve a stationary state with a longitudinal distance of its foremost point to the nearest edge of the stop line between 0 and 1,5 m or when no stop line is present 3m before the zebra crossing stripes | AV can continue if: VRU lateral distance 4m (going away), 7m (coming towards) from the AV footprint<br>AV longitudinal distance between 0 and 1.5 m of the stop line |

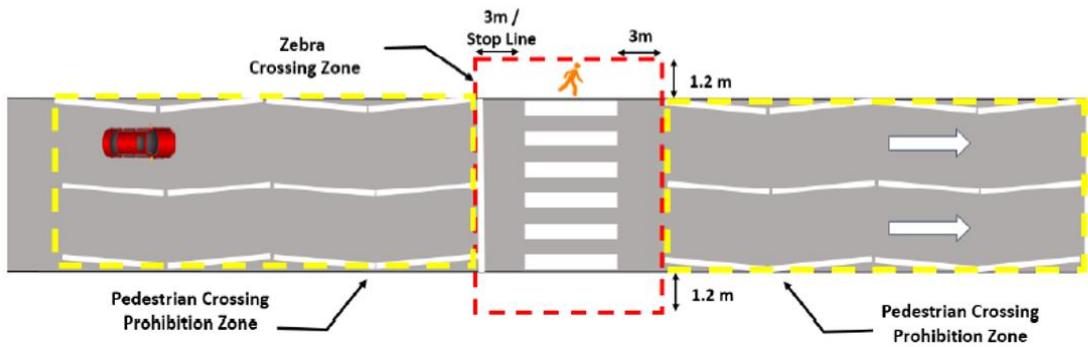

Figure 6: Zebra Crossing Zone as defined in TR68-1 [1].

At the left turn slip road, in this situation as with most left turn slip roads in Singapore, there is a zebra crossing. The current guidelines are seen in Table 39 above, with additional description of a zebra crossing zone from TR68-1 [1] in Figure 6 above. Most AV developers agree the current guidance is sufficient for expected behaviour at a zebra crossing. However as mentioned earlier, pedestrian intentions are a current challenge by developers. One question CETRAN posed was is it wrong for an AV to stop if pedestrians are further away or if there are no pedestrians. CETRAN shared that it would be ideal if an AV does not stop as it could potentially reduce traffic flow, but with safety as a priority, this could be deemed a minor non-conformity in AV assessment and could be considered in future implementation with further analysis.

### *4.2.2.2.6.    Stopping at Give-Way Line*

Table 40: Current guidelines from TR68-1, BTD and FTD where applicable at a slip road give-way line

| Slip road give way line | Parameters |
|---|---|
| **TR68-1: 6.1.h)** When merging into major roads from slip lanes, match speed of traffic on the major road for a smooth merge. | |
| **TR68-1: 6.4)** When stopping at a stop line, the AV should achieve a stationary state with a longitudinal distance of its foremost point to the nearest edge of the stop line between 0 m and 1.5 m. | 0 – 1.5m |
| **TR68-1: 7.10.1**) When approaching a non-signalised junction with a major road, an AV shall reduce its speed and give way to traffic on the major road. If there is a STOP sign, an AV shall stop at the stop line with the provisions of 6.4. | |
| **TR68-1: 7.10.2)** An AV shall comply with the "Give Way" rule at non-signalised junctions (BTD81)<br>**BTD81**: At an uncontrolled junction where there are no traffic lights, "GIVE WAY" sign or "STOP" sign: | |
| **BTD 80**: When approaching a junction with a major road, slow down gradually and give way to traffic on the major road. Where there is a "STOP" sign, stop before the stop-line. | |

At the end of a left turn slip road, there is a give way line. The current guidelines are seen in Table 40 above. One issue CETRAN identified was that at different locations in Singapore, the distance from the give-way line could be soon after a zebra crossing, which could result in a vehicle blocking the zebra crossing in part or fully depending on the length of the vehicle. CETRAN further identified that there is a possibility that AV would stop even if there is no traffic oncoming on the main stretch of road the AV is intending to merge into, and it could be contradictory to TR68-1 secondary directive of maintaining free movement of traffic [1]. The recommendation would be that an AV does not need to stop if there is no oncoming traffic. However, AV developers did feedback that they do program an AV to stop at a give-way line at the end of a left turn slip road even though there are no vehicles oncoming, to evaluate traffic further in the current maturity of the technology.

In view of that, one recommended additional guideline CETRAN would recommend to not block a zebra crossing would be:

1. If an AV vehicle length is longer than the available space between a zebra crossing and the give-way line, the vehicle should wait at the stop line before a zebra crossing, until it can safely join traffic coming from the right.

However, it is noted that if the AV stops at the stop line before the zebra crossing, it may not have a good view of traffic from the right. This is then treated in the next section on resolving occlusion to complete this recommended guideline.

### 4.2.2.2.7. *Resolving Occlusion*

Table 41: Current guidelines from TR68-1, BTD and FTD where applicable for occlusion situation

| Occlusion situation | Parameters |
|---|---|
| **TR68-1 6.1.c:** Speed should account for the road surface conditions and current and upcoming range of object sensor detection and visibility to other road users. When faced with areas of occlusions or range restrictions, an AV shall react to reasonable worst-case assumptions of actors being present just outside of field of view. In the event that occlusions will not be satisfactorily removed without forward motion, the AV should be allowed to creep forward at a reduced speed until the occlusions have been removed | |

This additional decision point on occlusion is included in this traffic situation as there have been instances where another vehicle such as the red TSV seen in Figure 2 could be a big vehicle and occlude the view of oncoming traffic. From the previous step of analysis, if the space between the give-way line and zebra crossing was small such that an AV would block the crossing while stopped at the give-way line, it is recommended to stop before the zebra crossing but this could result in an occluded by the other vehicle stopped at the traffic light junction. In such an instance, there is a guidance that allows for an AV to creep forward until the occlusion has been removed as seen in Table 41above.

For such an instance, CETRAN suggest that it would be acceptable for an AV to creep forward onto the zebra crossing if there are no pedestrians crossing, even if the vehicle is longer than the space available and will result in blocking the zebra crossing until it is safe to complete the left turn.

## 4.2.3 Summary of recommended additional guidelines in this example

Several recommended additional guidance was done through the detailed analysis of guidelines pertaining to the traffic situation described in this section 4.2.1. The recommended additional guidance are:

1) Utilize a safe following distance of "so that if the leading vehicle would initiate a complete stop, the AV would be able to stop behind that vehicle with a longitudinal distance of at least 2 m" in TR68-1 6.3.4 to be used for assessment.
2) Complete overtaking manoeuvre with at least 3 seconds before start of left turn slip road.
3) Safe distance for lane change to be:

a) When changing lanes, the AV should be maintaining a safe distance of 1 car length per 16km/h of the TSV speed or 9 seconds TTC whichever is bigger to the TSV/cyclist coming from behind in the adjacent lane (i.e. when the AV starts to cross the lane marking)
b) If there is an oncoming vehicle in the adjacent lane during the overtaking manoeuvre, the AV should change back into its original lane with at least a 2 second time gap (time to collision) with the TSV/cyclist oncoming in the adjacent lane.
c) A recommended value for lateral acceleration should be used in TR68-1, such as UNECE 79 5.6.2.1.3 [8] according to vehicle class and vehicle speeds, such that bad behaviour of 'sudden cut ins' can be better defined, together with safe distances to other actors. This is recommended to follow a maximum of 3m/s$^2$ for speeds between 10-60km/h (M1, N1 vehicle category) and with a moving average over half a second of the lateral jerk generated to not exceed 5m/s$^3$ [8].

4) When entering a left turn slip road with a zebra crossing approaching a traffic junction: An AV should start turning at 3metre after the start of the geometric start of slip road, maintaining a smooth trajectory from the centre of the original lane into the centre of the left turn slip road lane, so as not to encourage motorcyclists from entering from the left of the AV. (This is applicable for urban roads and not for exits from highways).
5) If an AV vehicle length is longer than the available space between a zebra crossing and the give-way line, the vehicle should wait at the stop line before a zebra crossing, until it can safely join traffic coming from the right. If however, an AV's view of oncoming traffic is occluded when stopped at the zebra crossing, it can creep forward when there are no pedestrians crowing and block the zebra crossing until it is safe to complete the left turn.

## 4.3 Example 2- Discretionary right turn at traffic junction

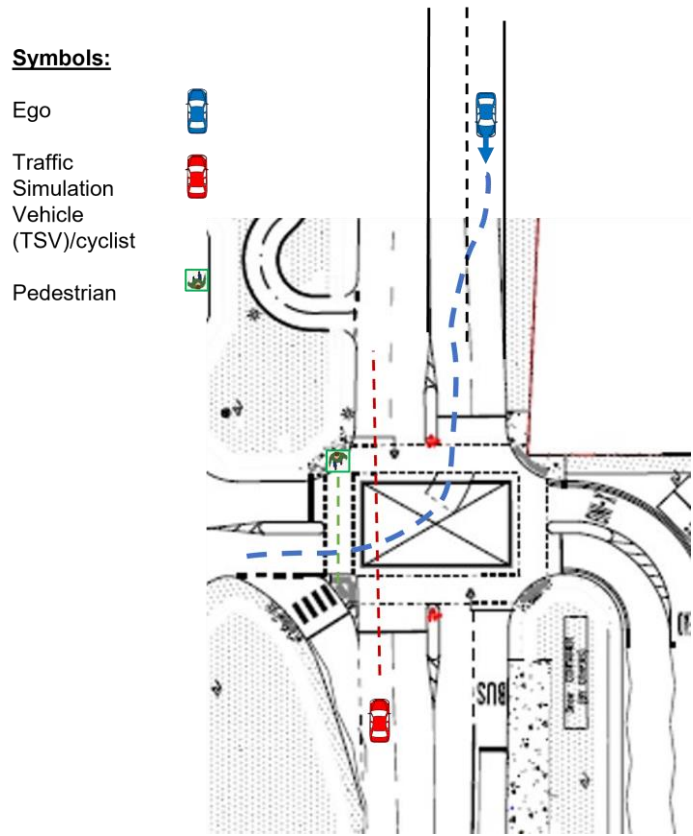

Figure 7: Discretionary right turn where an AV (ego vehicle) is approaching a traffic junction and planning to execute a discretionary right turn, with other actors as depicted.

### 4.3.1 General Situation

**Description**

- Ego vehicle /AV plans to make a discretionary right turn at a traffic light junction.
- An actor vehicle is travelling in the opposite traffic direction towards the junction.
- A pedestrian (with green pedestrian light) is crossing the road where the Ego is intending to turn into

**Assumptions:**

- Oncoming vehicle and pedestrian at crossing as 2 other actors
- Traffic light is green light (not green right arrow) for ego (discretionary right turn)
- Green light for TSV and pedestrian.

**Missing guidance**

- How to judge what is a safe distance - missing specific metrics in guideline.

### 4.3.2 Analysis of situation

#### *4.3.2.1 Decision Steps Required*

1. **Lane change into right turning lane**
    - When should I start lane change manoeuvre?
    - Is traffic in adjacent lane clear?
    - How much longitudinal gap do I need to keep to traffic in adjacent lane.
    - Slowing down in current lane, to lane change.
2. **Slow down at traffic junction**
    - When to start slowing down? Check lights, check traffic ahead
3. **Move into stop pocket**
    - When to move into pocket? Is traffic light green to move ahead?
4. **Judging safe distance to move off**
    - Oncoming Vehicle – sufficient time buffer to clear AV trajectory before time to collision?
    - Pedestrian crossing – sufficient distance to clear AV trajectory before time to collision?
5. **Executing turn**
    - What speed is necessary for turn curvature
6. **Entering lane**
    - Which lane to turn right into?

#### *4.3.2.2 Guidelines of each decision step*

##### *4.3.2.2.1. Lane change into right-turning lane*

Within this first decision step, there are a couple of actions an AV would need to carry out: 1. Signal, 2. Adjust speed, 3. Smoothly and gently steer into the lane intended, 4. Cancel signal and resume speed.

Table 42: Current guidelines from TR68-1, BTD and FTD where applicable on lane change

| Steps within Lane Change | Current guidelines |
|---|---|
| Signal | **TR68-1: 7.5** If the AV plans a lane change in the absence of an immediate safety concern, the lane change manoeuvre shall commence not before **3seconds** have passed after the activation of the signal indicator to the intended side of lane change as defined in 5.6.4.6.4 of ECE/TRANS/WP.29/GRRF/85 and **BTD131.** <br> **FTD 106a) (i)** Check mirrors; <br> (ii) Signal your intention; |
| Adjust speed | **TR68-1: 7.5a)** Before commencing a lane change manoeuvre, the AV shall ensure that the manoeuvre does not force any ORU to take potentially unsafe evasive action. <br> **FTD 106b) (i)** Adjust your speed; <br> (ii) You may have to slow down or speed up depending on the traffic situation behind you. |
| Smoothly and steer gently | **FTD 106c)** When it is safe, accelerate smoothly and steer gently into the lane intended without interrupting the flow to traffic |
| Cancel signal, resume speed | **TR68-1: 7.5c)** The AV shall cancel its signal after the lane change manoeuvre is completed, i.e. its footprint is enclosed by the boundaries of the new lane. <br> **FTD 106d)** Cancel your signal and resume your normal speed. |

Table 42 above shows the current guidelines on the steps required for an AV to carry out when deciding to lane change. There appears to be ambiguity in what constitutes as safe to enter the adjacent as in the earlier example of performing a lane change for overtaking, and what is gentle steering is. CETRAN has had experience where an AV performs an abrupt lane change, as is the experience from human drivers. Hence the additional recommended guidelines for lane change would follow that described in Section 4.2.2.2.3 above. In this traffic situation, an AV would perform a lane change before proceeding to perform a right turn, to cancel the right turn signal after the lane change and turn it back on might seem slightly contradictory. There is an additional recommendation for this included in the overall additional recommended guidelines in this step.

1. When changing lanes, the AV should be maintaining a safe distance of 1 car length per 16km/h of the TSV/cyclist speed or 9 seconds TTC whichever is bigger to the TSV/cyclist coming from behind in the adjacent lane (i.e. when the AV starts to cross the lane marking)
2. A recommended value for lateral acceleration should be used in TR68-1, such as UNECE 79 5.6.2.1.3 [8] according to vehicle class and vehicle speeds, such that bad behaviour of 'sudden cut ins' can be better defined, together with safe distances to other actors. This is recommended to follow a maximum of $3m/s^2$ for speeds between 10-60km/h (M1, N1 vehicle category) and with a moving average over half a second of the lateral jerk generated to not exceed $5m/s^3$ [8].
3. If there is overlap to AV's next action-another right turn within 10s, continue signal, otherwise cancel signal.

### 4.3.2.2.2. *Slow down at traffic junction*

In the next step of approaching a traffic junction, the following guidelines in Table 43 below applies. The guidelines are clear, and no missing or conflicting guidelines are observed for this decision step.

Table 43: Current guidelines from TR68-1, BTD and FTD where applicable on slowing down at traffic junction, signalized junction from 'Traffic Lights- Nominal (3.2.2.1.1)' and 'TURNS-turning right (3.2.2.1.6)'

| Steps | Current guidelines |
|---|---|
| Slow down | **TR68-1: 7.2.6d iii) (FTD 118)** be at a suitable speed in readiness for any planned manoeuvres at the junction **(eg. Turning left or right)**<br>**FTD118:** Do not accelerate when you are approaching a signalized junction. Always be prepared for the signal light to change by reducing your speed. This will give you time to decelerate evenly when the traffic light changes from green to amber. |
| Signal right | **TR68-1: 7.10.3 b)** When turning right going straight, into a shared lane for turning right and going straight, signal early so that straight-going vehicles from behind can take an alternate lane to proceed **(FTD116).**<br>**FTD 116:** When turning right into a 'shared' lane for turning right and going straight, signal early so that straight going vehicles from behind can take an alternate lane to proceed. |

#### 4.3.2.2.3. *Move into right turning pocket*

Table 44: Current guidelines from TR68-1, BTD and FTD where applicable on moving into a right turning pocket from 'Turning right – discretionary right turn' (3.2.2.1.7)

| Steps | Current guidelines |
|---|---|
| Move into right turning pocket | **TR68-1: 7.10.3 a**) When turning right at a signalised junction, an AV shall first stop at the right turning pocket. Otherwise, where there is no right turning pocket, an AV shall first stop at a safe distance away from opposing traffic. The AV shall only proceed to make the right turn when there are neither oncoming vehicles travelling in the opposite direction nor pedestrians, cyclists and PMD riders crossing the road on its right. **BTD 54b**) Vehicles turning right at road junctions should stay within the pocket until it is clear to complete the turn |
| Stopping at stop line | 6.4) When stopping at a stop line, the AV should achieve a stationary state with a longitudinal distance of its foremost point to the nearest edge of the stop line **between 0 m and 1.5 m.** |

In this step of approaching a right turning pocket at a traffic junction, the guidelines in Table 44 above applies. The guidelines are clear, and no missing or conflicting guidelines are observed for this decision step.

#### 4.3.2.2.4. *Judging safe distance to move off*

Table 45: Current guidelines from TR68-1, BTD and FTD where applicable on judging safe distance to move off from right turning pocket, with guidelines from 'Traffic lights-nominal' (3.2.2.1.1), 'TURNS-turning right' (3.2.2.1.6), and 'Pedestrian Crossing' (3.2.2.4).

| Steps | Current guidelines |
|---|---|
| Judging safe distance to move off | **TR68-1: 7.2.6e**) The AV should not proceed past the stop line if there is no space for exiting the junction (e.g. due to queued vehicles **(FTD120)** |
| 1. Oncoming Vehicle | **FTD113 (3)** <br> At 'f' Drive slowly and cautiously towards the centre of the junction; Give way to oncoming vehicles; Wait until it is safe to cross or wait for the green arrow signal to appear; Turn swiftly to the correct lane keeping a look-out for pedestrians crossing at the junction. |
| 2. Pedestrian | **TR68-1: 7.2.6c**) Give way to pedestrians crossing the road **(BTD145:** When approaching a pedestrian crossing, ALWAYS –(a)be ready to slow down or stop so as to give way to pedestrians. <br> (b)signal to other drivers your intention to slow down or stop; <br> (c)allow yourself more time to stop when the road is wet. <br> **TR 68 - 7.9.4** Stopping at Zebra crossing. <br> AV shall stop outside the zebra crossing zone when VRU are present inside the zebra crossing zone with one or more of the following attributes: <br> a) On the road surface less than 4 m away laterally from the AV footprint <br> b) Moving towards the AV's path and is within 7m laterally from the AV footprint <br> c)Has displayed the intent to cross the AV's path and is within 7m laterally from the AV 's footprint. <br> AV shall stop outside the zebra crossing zone at the stop line in accordance with **TR 68 - 6.4** or stopping at stop lines. <br> When stopping at stop lines the AV should achieve a stationary state with a longitudinal distance of its foremost point to the nearest edge of the stop line |

| | between 0 and 1,5 m or when no stop line is present 3m before the zebra crossing stripes |

In the next step of approaching a traffic junction, the following guidelines in Table 45 above applies. There is ambiguity in what is a safe distance for completing the right turn in the guidelines and has resulted in some experience where AVs took a risk in completing turn where a tester felt unsafe, although no crash was observed. Perceived safety is key for surrounding traffic and also for assessors when conducting assessments. Discretionary right turns were some of the traffic situations where fatal accidents have occurred and have sparked "public safety concern" [9]. As noted by the Land Transport Authority (LTA) in establishing the need for removing discretionary right turn junctions, such turns require vehicles to "assess the flow of oncoming traffic and pedestrians crossing before turning right once the coast was clear" [10]. While such infrastructure is being largely reduced [9], CETRAN is aware AVs would face the possibility of making such discretionary right turns when conducting trial on Singapore road, and hence an additional guidance for determining safety is necessary.

CETRAN proposes that a metric to assess safety to carry out the right turn:

> *An AV should cross the path of an oncoming vehicle in a discretionary right turn if it is able to complete the crossing with at least a 2 second time buffer to the oncoming TSV/cyclist at the end of the turn, out of the traffic junction.*

This was trialled on CETRAN's test track and allowed assessors to experience to judge after the right turn was executed as shown in Figure 8 below. An AV would likely need to determine the required safe distance that the oncoming vehicle needs to be away from its intended path before it executes the turn. However, trials on CETRAN test track proved that human assessors find it difficult to judge this safe distance at the start of the turn, and a metric at the end of the turn would be easier to implement. The following recommended guidance would guide developers on what is the acceptable result and allow them flexibility in implementing the necessary algorithm and planning behaviour for an AV to execute a discretionary right turn.

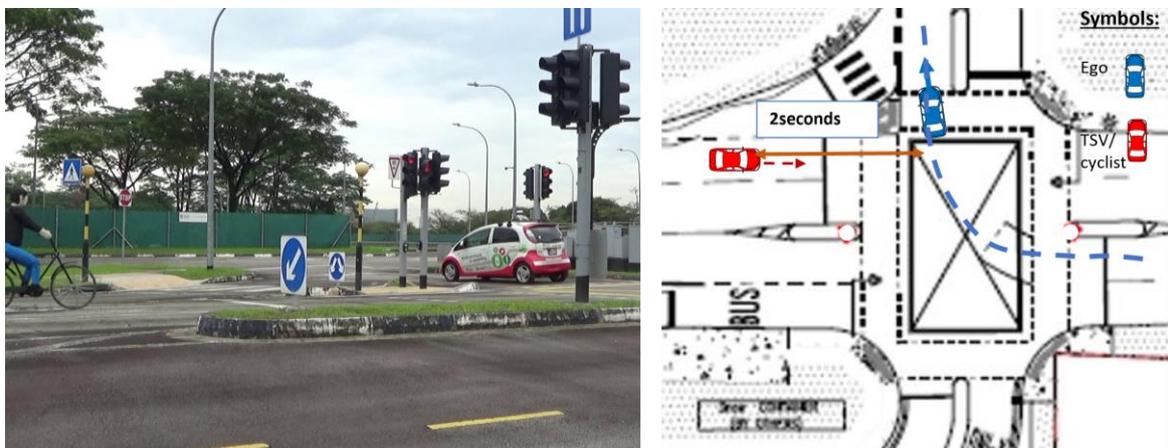

Figure 8: Test on CETRAN track for a 2seconds TTC as a guideline for judging safe distance to carry out a discretionary right turn.

An AV also needs to judge if its path is clear with a pedestrian crossing likely to be showing green light at the other end of a discretionary right turn. The current guideline for pedestrian crossing is TR 68-1 7.2.6c, however there are no values of lateral clearance to keep to pedestrians. CETRAN recommends applying the lateral clearances necessary from TR68-1 7.9.4 from zebra crossing (which is a) On the road surface less than 4 m away laterally from the AV footprint; b) Moving towards the AV's path and is within 7m laterally from the AV footprint; c)Has displayed the intent to cross the AV's path and is within 7m laterally from the AV 's footprint. ) to be applied to a pedestrian crossing at a traffic junction as well.

Hence there are two additional recommended guidelines from this step of this traffic situation:

1. An AV should cross the path of an oncoming vehicle in a discretionary right turn if it is able to complete the crossing with at least a 2 second time[2] buffer to the oncoming TSV/cyclist at the end of the turn, out of the traffic junction.
2. Applying the lateral clearances necessary from TR68-1 7.9.4 from zebra crossing to be applied to a pedestrian crossing at a traffic junction as well.

### 4.3.2.2.5. *Executing turn*

Table 46: Current guidelines from TR68-1, BTD and FTD where applicable on executing turns (e.g. from ride comfort)

| Steps | Current guidelines |
|---|---|
| Executing turn | **BTD199/ FTD 197:** The greater the travelling speed around the curve or the sharper the curve, the more the vehicle will be pushed from its path. You should therefore reduce speed when going round a bend. The diagram on the right shows the appropriate speed and the dangerous speed for each turning radius. |

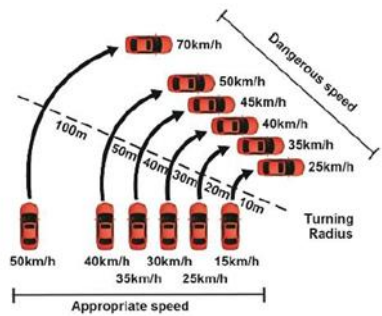

Figure 9: Diagram from BTD and FTD on appropriate speeds for turning [2] [3]

When the AV is executing the right turn across the junction, it should follow the guideline in Table 46 above. This is linked to the appropriate lateral acceleration during the turn, where UNECE 79 provides further recommendation that the lateral acceleration shall not exceed "$1m/s^2$ in addition to the lateral acceleration generated by the lane curvature" [8].

### 4.3.2.2.6. *Entering lane*

Table 47: Current guidelines from TR68-1, BTD and FTD where applicable on entering lane after a turn, from 'TURNS-turning right, (3.2.2.1.6)

| Steps | Current guidelines |
|---|---|
| Entering lane | **BTD84)** When turning right into a two-way street, turn into the lane just left of the centre line of the road you are turning into. **BTD85)** When turning right into a one-way street, turn into the extreme right lane of the road you are turning into. |

After the AV has executed the right turn across the junction, it would need to enter the appropriate lane and the guidelines in Table 47 above applies. This maybe a source of confusion for AVs if in case there are two right turning lanes turning into a three-lane road. CETRAN recommends an additional guideline:

1. If an AV is in the left lane of the 2 lanes turning, it should turn into the 2nd lane from the left of the central divider (e.g. middle lane of 3 lanes).

### 4.3.3 Summary of recommended additional guidelines from this example

Several recommended additional guidance was raised through the detailed analysis of guidelines pertaining to the traffic situation described in this section 4.3. The recommended additional guidance are:

1) When changing lanes, the AV should be maintaining a safe distance of 1 car length per 16km/h of the TSV/cyclist speed or 9 seconds TTC whichever is bigger to the TSV/cyclist coming from behind in the adjacent lane (i.e. when the AV starts to cross the lane marking)
2) A recommended value for lateral acceleration should be used in TR68-1, such as UNECE 79 5.6.2.1.3 [8] according to vehicle class and vehicle speeds, such that bad behaviour of 'sudden cut ins' can be better defined, together with safe distances to other actors. This is recommended to follow a maximum of $3m/s^2$ for speeds between 10-60km/h (M1, N1 vehicle category) and with a moving average over half a second of the lateral jerk generated to not exceed $5m/s^3$ [8].
3) If there is overlap to AV's next action-another right turn within 10s, continue signal, otherwise cancel signal.
4) An AV should cross the path of an oncoming vehicle in a discretionary right turn if it is able to complete the crossing with at least 2 second time buffer to the oncoming TSV/cyclist at the end of the turn, out of the traffic junction.
5) Applying the lateral clearances necessary from TR68-1 7.9.4 from zebra crossing to be applied to a pedestrian crossing at a traffic junction as well.
6) If an AV is in the left lane of the 2 lanes turning, it should turn into the 2nd lane from the left of the central divider (e.g. middle lane of 3 lanes).

## 4.4 Example 3- Overtaking parked vehicle

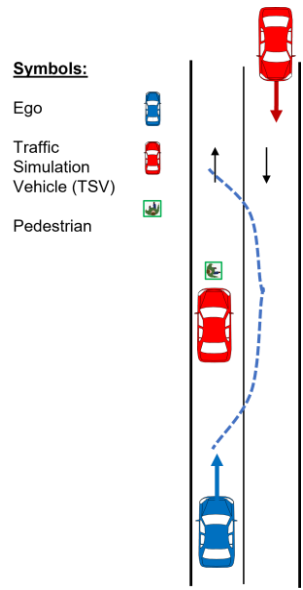

Figure 10: An AV (ego vehicle) approaching a parked vehicle and planning to execute an overtaking manoeuvre.

### 4.4.1 General Situation

**Description:**

- Ego vehicle /AV is blocked by an illegally parked vehicle in its lane ahead, plans to carry out an overtaking manoeuvre.

**Assumptions:**

- Stationary car implies an occlusion lies ahead

**Missing guidance**

- How to judge what is a safe distance when changing lanes- missing specific metrics in guideline.
- Quality of lane change – what counts as abruptness?
- How long can a vehicle stay on adjacent lane when carrying out overtaking

### 4.4.2 Analysis of situation

#### 4.4.2.1 Decision Steps Required

1. **Maintain safety and lane discipline when approaching stopped vehicle**
    - What is a safe distance to maintain to the vehicle ahead, to start slowing down
2. **Judging safe conditions to start overtaking**
    - Is traffic in adjacent lane clear?
    - When is a safe distance to oncoming traffic or vehicle from behind in adjacent lane
3. **Overtaking: Lane change and overtaking procedure**
    - What is a smooth lane change
    - **While in adjacent lane**
    - To continue look out for further oncoming traffic

- Lateral distance to maintain to parked vehicle
- Possible pedestrian in occlusion zone of parked vehicle
4. **When to pull back into original lane – left lane change**
    - What is a good distance to pull in ahead of parked vehicle

### 4.4.2.2    *Guidelines of each decision step*

#### 4.4.2.2.1.    *Approaching parked vehicle*

When approaching an illegally parked vehicle ahead, the following guidelines exist as seen in Table 48 below.

Table 48: Current guidelines from TR68-1, BTD and FTD where applicable on lane discipline and stop distances to vehicle ahead

| Steps | Current guidelines |
|---|---|
| Lane Discipline | **TR68-1:** 7.4a) The AV shall obey directions indicated by arrows marked in lanes **(BTD64).**<br>**TR68-1:** 7.4d) The AV shall keep within its lane unless it is performing a lane change or overtaking manoeuvre **(FTD311, 160b).**<br>**BTD 160c), FTD312** Do not weave in and out of traffic lanes as it would cause confusion and danger to others.<br>**FTD108**: Always keep well to the left when driving along two-way streets or dual carriageways, unless your path of travel is obstructed by road works, parked vehicles, etc. In such circumstance, you may cross the centre line or move to the lane on your right. On doing so, take care to ensure that your intended path is safe and clear before you move to the right. |
| Maintain safe distance to vehicle ahead | **TR68-1:** 6.3.4) The AV shall maintain a safe car following distance (BTD 135) to all moving vehicles in the same lane "ahead" with "established lane presence". A vehicle will be considered as "ahead" in lane if their front bumper is further along the lane in a longitudinal sense than the AV's front bumper. A vehicle will be considered to have "established lane presence" if the majority of their footprint is within the lane boundaries.<br>The safe following distance shall be determined so that if the leading vehicle would initiate a complete stop, the AV would be able to stop behind that vehicle with a longitudinal distance of at least 2 m. The AV should respond proactively when other vehicles are observed to be cutting into its lane so as to prevent or minimise violation the safe following distance as much as possible. |

The guidelines are generally clear on what is expected when approaching a parked vehicle. Lane discipline is required prior to decision of lane change, by following the lane marking direction and keeping within lane. The guideline on following distance could be used for what is appropriate distance of 2 metres to stop if need behind the stationary vehicle. It could be mentioned that a metric should be provided for what constitutes weaving. From CETRAN's experience, this could be stated as a vehicle moving left and right momentarily by more than 0.5 metres, but further definition of time would be investigated in the future.

#### *4.4.2.2.2.  Judging safe conditions to carry out overtaking*

In deciding on overtaking the parked vehicle, some considerations are to check for traffic in adjacent lane. The relevant guidelines have been selected in Table 49 below. An AV would need to adjust its speed in lane, to slow down and stop if there is traffic in adjacent lane and unable to pull out. Crossing a double white line is permissible if the lane is blocked by an illegally parked vehicle.

Table 49: Current guidelines from TR68-1, BTD, and FTD where applicable on judging safe conditions for overtaking, from 'Overtaking-General'

| Steps | Current guidelines |
|---|---|
| Judging safe conditions for overtaking | **TR68-1 7.6 a)** An AV shall only perform an overtaking manoeuvre when it is safe to do so.<br>**TR68-1 7.6 b)** An AV shall check for any traffic signs prohibiting overtaking before initiating an overtaking manoeuvre.<br>**TR68-1: 7.5** If the AV plans a lane change in the absence of an immediate safety concern, the lane change manoeuvre shall commence not before 3seconds have passed after the activation of the signal indicator to the intended side of lane change as defined in 5.6.4.6.4 of ECE/TRANS/WP.29/GRRF/85 and **BTD 131**. The lane change manoeuvre starts when the AV stars to move laterally so that it gets closer to the lane to which it intended to change. The lateral movement to approach the lane marking and the lateral movement necessary to complete the lane change manoeuvre should be completed as one continuous movement.<br>**TR68-1 7.6 c) ii)** When performing an overtaking manoeuvre, an AV shall never cross the double white line, except when its path of travel is obstructed by parked vehicles.<br>**TR68-1: 7.5a)** Before commencing a lane change manoeuvre, the AV shall ensure that the manoeuvre does not force any ORU to take potentially unsafe evasive action. |

As it is unclear on how to judge safety when overtaking, CETRAN recommends an additional guideline that was utilized for lane change previously. These will help an AV decide if it has sufficient safety to other traffic users for its overtaking manoeuvre, and this will be used to judge the safety of its behaviour by assessors as well when the overtaking manoeuvre begins.

1) If there is an oncoming vehicle in the adjacent during the overtaking manoeuvre, the AV should change back into its original lane with a 2 second time gap (time to collision) with the TSV/cyclist oncoming in the adjacent lane.
2) When changing lanes, the AV should be maintaining a safe distance of 1 car length per 16km/h of the TSV speed or 9 seconds TTC whichever is bigger to the TSV/cyclist coming from behind in the adjacent lane (i.e. when the AV starts to cross the lane marking)

Copyright©2023 Nanyang Technological University

### 4.4.2.2.3. _Overtaking parked vehicle: Lane Change and Overtaking Procedure_

When an AV has decided on overtaking the parked vehicle in a similar traffic situation, it should follow the guidelines that are found in Table 50 below when carrying out a lane change action, and for overtaking procedure as a lane change is required when overtaking. When the AV is in the adjacent lane, there are also guidelines to take note of.

Table 50: Current guidelines from TR68-1, BTD, and FTD where applicable on lane change and overtaking.

| Steps | Current guidelines |
|---|---|
| Signal Right | **TR68-1: 7.5** If the AV plans a lane change in the absence of an immediate safety concern, the lane change manoeuvre shall commence not before 3seconds have passed after the activation of the signal indicator to the intended side of lane change as defined in 5.6.4.6.4 of ECE/TRANS/WP.29/GRRF/85 and **BTD 131**. The lane change manoeuvre stars when the AV stars to move laterally so that it gets closer to the lane to which it intended to change. The lateral movement to approach the lane marking and the lateral movement necessary to complete the lane change manoeuvre should be completed as one continuous movement. **BTD131:** Always give clear signals well in advance of your intentions (at least 3 seconds) before your manoeuvre so that other road users can interact safely |
| Lane Change/ Overtaking decision | **TR68-1: 7.5a)** Before commencing a lane change manoeuvre, the AV shall ensure that the manoeuvre does not force any ORU to take potentially unsafe evasive action.<br>**TR68-1 7.6 d)** An AV shall only move into the lane for oncoming traffic when performing an overtaking manoeuvre if the intended path is safe and clear of traffic<br>**FTD 136** When overtaking you have to estimate not only the space ahead of you but also the speed and distance of oncoming vehicles |
| Lane Change/ Overtaking action | **TR68-1 7.6 e)** An AV shall adopt the following steps when performing an overtaking manoeuvre:<br>Signal right, Right lane change, Accelerate, Signal left, Left lane change, Cancel signal, Resume normal speed<br>**FTD 106c)** When it is safe, accelerate smoothly and steer gently into the lane intended without interrupting the flow to traffic. |
| Overtaking in adjacent lane | **TR68-1 6.3.2.c)** Whenever the AV is in motion with speed less than 30 km/h, the lateral clearance to pedestrians on the road surface and **not laterally moving away** from the AV's path shall be no less than 1 m.<br>**TR68-1 6.3.2.d)** For AV motion at speeds equal to or above 30 km/h, the lateral clearance to pedestrians on the road surface and **not laterally moving away** from the AV's path shall be no less than 1.5 m.<br>**FTD 221:** When passing by a parked vehicle, look out for drivers or passengers opening the doors of their vehicles and keep a safe gap of about 1 meter between you and the parked vehicles.<br>**BTD 147**: Watch for pedestrians who come out suddenly from behind stationary vehicles and other obstructions. Be very careful near schools and bus stops. |

Overall, the guidelines are clear on what is required for overtaking procedure in TR68-1 7.6 e). First there is need to indicate for 3 seconds, then perform the right lane change by gently and smoothly steering.

However, there are parameters in TR68-1 7.5a) and b) on what constitutes as 'safe distance' that will not force ORU to take evasive action and a safe following distance. This addressed in the previous decision step with CETRAN's recommended guideline for judging safe distance. When conducting a lane change, it is not clear what is 'smooth and gentle'. As addressed in an earlier example, CETRAN recommends the additional guideline for lateral acceleration:

1. A recommended value for lateral acceleration should be used in TR68-1, such as UNECE 79 5.6.2.1.3 [8] according to vehicle class and vehicle speeds, such that bad behaviour of 'sudden cut ins' can be better defined, together with safe distances to other actors. This is recommended to follow a maximum of 3m/s$^2$ for speeds between 10-60km/h and with a moving average over half a second of the lateral jerk generated to not exceed 5m/s$^3$ [8].

The above guidelines in Table 50 also apply to the AV while it is executing its overtaking of a parked vehicle, with prescribed lateral clearance to pedestrians in TR68-1 6.3.2 c and d. There is a potential for pedestrians on the road past the parked vehicle and occluded from the parked vehicle itself. It is also expected there is a 1metre lateral clearance maintained to the parked vehicle as stated in FTD211 below. However, in CETRAN's experience, to maintain to a 1metre distance and the potential occlusion of pedestrians behind the parked vehicle, the AV should proceed with caution. CETRAN thus recommend this additional guideline when passing a parked vehicle:

1. An AV should show caution when overtaking a parked vehicle. To proceed with a lateral clearance of 1metre, it should slow down and pass with a maximum speed of 30km/h to the parked vehicle. This reduced speed should allow the AV to have sufficient braking to avoid collision.

### 4.4.2.2.4. *Lane change back into original lane*

After passing the parked vehicle, the AV should carry out a left lane change to return back to its original lane. The guidelines for lane change still applies as seen in Table 50 above. One earlier recommended guideline CETRAN provided is that there should be at least a 2-seconds time-gap to oncoming vehicles in adjacent lane after the AV has returned into its original lane. There is however no guideline on how far ahead of the parked vehicle is considered a safe distance to pull back in. In the absence of oncoming vehicles, it is also unclear how long an AV can remain in the other lane (if the road has multiple lanes in the same direction) before a lane change back.

CETRAN recommends the following additional guideline:

1. An AV should return to its original lane if it is clear of traffic, after overtaking the parked vehicle after it has at least 1- 1.5x ego vehicle's car length ahead of the parked vehicle and with a maximum of 20 seconds past the parked vehicle in the other lane.

### 4.4.3 Summary of recommended additional guidelines from this example

Several recommended additional guidance was done through the detailed analysis of guidelines pertaining to the traffic situation described in this section 4.4. The recommended additional guidance are:

1) Safe distance for lane change to be:

    a) When changing lanes the AV should be maintaining a safe distance of 1 car length per 16km/h of the TSV speed or 9 seconds TTC whichever is bigger to the TSV/cyclist coming from behind in the adjacent lane (i.e. when the AV starts to cross the lane marking)
    b) If there is an oncoming vehicle in the adjacent during the overtaking manoeuvre, the AV should change back into its original lane with at least a 2 second time gap (time to collision) with the TSV/cyclist oncoming in the adjacent lane.

2) A recommended value for lateral acceleration should be used in TR68-1, such as UNECE 79 5.6.2.1.3 [8] according to vehicle class and vehicle speeds, such that bad behaviour of 'sudden cut ins' can be better defined, together with safe distances to other actors. This is recommended to follow a maximum of $3m/s^2$ for speeds between 10-60km/h and with a moving average over half a second of the lateral jerk generated to not exceed $5m/s^3$ [8].

3) An AV should show caution when overtaking a parked vehicle. To proceed with a lateral clearance of 1metre, it should slow down and pass with a maximum speed of 30km/h to the parked vehicle.

4) An AV should return to its original lane if it is clear of traffic, after overtaking the parked vehicle after it has at least 1- 1.5x ego vehicle's car length ahead of the parked vehicle and with a maximum of 20 seconds past the parked vehicle in the other lane.

## 4.5 Example- Situations studied in collaboration with STEAS

In collaboration between STEAS and CETRAN, STEAS shared certain traffic situations where they have faced challenges to respect current guidelines. The following few examples encompass the issues they have shared, which CETRAN and STEAS then discussed and have come up with additional recommended guidelines for AVs on road.

### 4.5.1 Wide vehicle in adjacent lane

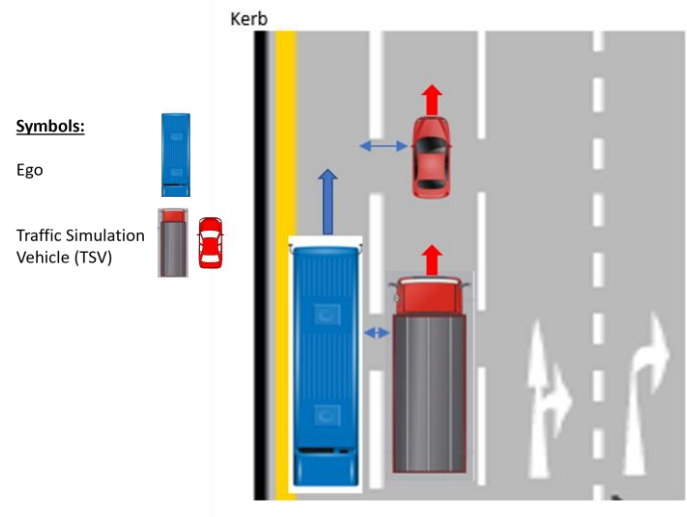

Figure 11: Traffic situation with slow vehicle travelling slow in adjacent lane.

#### 4.5.1.1 General Situation

**Description**

- Ego driving straight with slow moving vehicle (TSV) in adjacent lane.

**Assumptions**

- Ego moving along straight road in lane.
- TSV is slower than AV in adjacent lane(<25km/h).
- Multiple lanes on road

**Difficulty of situation; Missing Guidance**

- The lateral clearance to the moving TSV indicated in the FTD 227 is 1.5m. Maintaining this clearance can be a challenge due to the larger vehicle width for Class 4 (may be experienced by Class 3 vehicles as well).
- AV decision to continue moving in current lane by reducing lateral clearance to TSV or to slow down and stay behind adjacent TSV (respecting lateral clearance guideline).
- The concept of drivable area is not defined.

#### 4.5.1.2 Analysis

In this first traffic situation within the collaboration with STEAS, it was noted to be difficult to maintain the lateral clearance to a wide vehicle in an adjacent lane as stated in FTD, with Singapore's standard road width to be 3.2 to 3.5metres [11]. It was also further noted that the AV was expected to keep in the centre of the lane, however the definition of a driveable area in lane was not clearly defined. Hence through consultation with former Traffic Police members within CETRAN and STEAS teams, the following definition was defined as a first additional recommended guideline in Figure 12 below.

This clarifies that it is acceptable that a vehicle can bias to the left within the drivable area of the lane and drive over the kerbside marking in this specific case. For this traffic situation, there is only the current guideline of FTD 221 and 227 regarding lateral clearance to observe to vehicles based on their speeds [3]. However, a typical wide vehicle is 2.5metres in width, to maintain a 1.5metre clearance to a wide vehicle in the adjacent lane, with a standard lane width of 3.2-3.5 metres, both the Ego and TSV in adjacent lane would need to be at the extreme bias of their lanes which is not realistic and conflicts with the consensus to keep within centre of the lane. It was discussed that the AV can choose to stay behind the TSV in the adjacent lane if lateral clearance to move past is below required as stated in FTD. It was also proposed that a smaller lateral clearance is possible to allow for smooth traffic flow, but only if the ego moves at a reasonable speed relative the slow moving TSV in the adjacent lane. The following are the additional recommendation for this traffic situation.

The reduced clearance of 0.5metres to a slow moving TSV in the adjacent lane was recommended when the AV takes care to reduce its speed relative to the TSV, with an appropriate speed such as 20% higher than the TSV. In case the TSV is stationary in traffic, the AV can pass at a maximum of 30km/h at this reduced clearance of 0.5m. This was tested out on the CETRAN track as shown in Figure 13 below. As the lane width on CETRAN track are wider, a safety barrier was placed to indicate kerb location with a distance between the barrier and the test dummy vehicle as 3.7metre distance, which is a possible situation on a standard 3.2metre lane width location.

Some other traffic situations discussed with STEAS utilize this same recommended behaviour of lane discipline and are included in the appendix section 2 on situations regarding lane discipline.

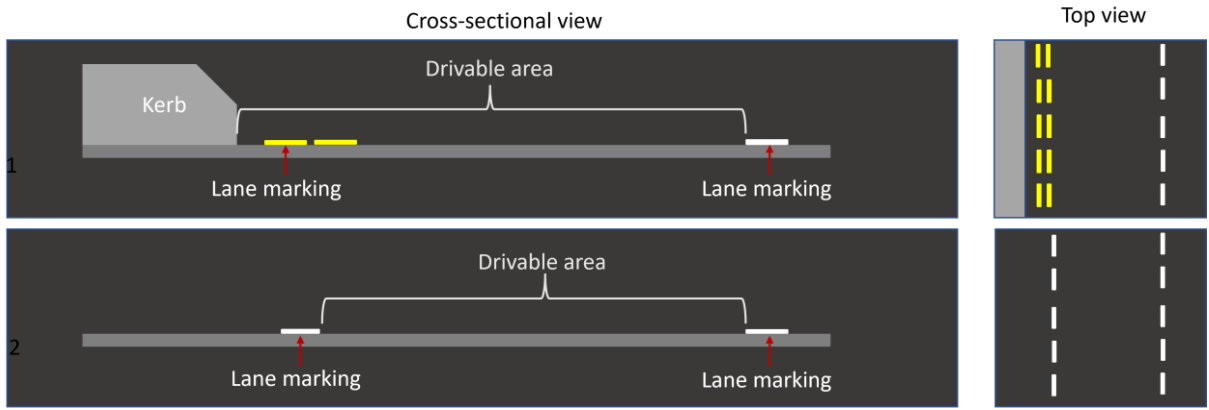

Figure 12: Additional recommended guideline of driveable area within lane for vehicles on road by STEAS and CETRAN.

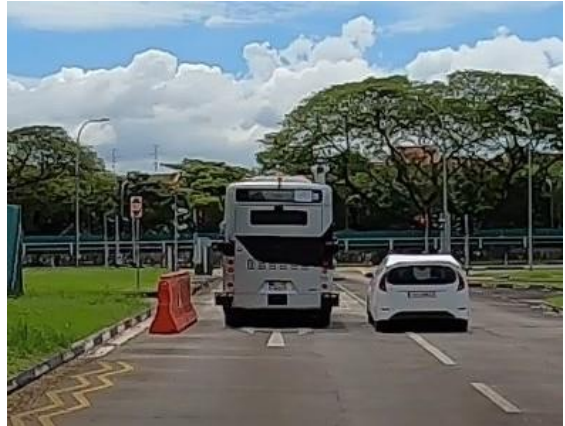
Figure 13: Test on CETRAN track with STEAS Class 4 vehicle next to a test dummy vehicle.

These are the recommended behaviour steps for such a situation:

1. AV should drive in the centre of the drivable area within a lane (as seen in Figure 12 above), it can also bias to the left of the lane to respect 1.5m in FTD 227 to a moving vehicle in the adjacent lane.
2. An AV can stay behind a slow TSV in the adjacent lane if the lateral clearance is less than the 1.5m in FTD 227 (see Figure 14 below).
3. The AV is allowed to drive pass the slow moving TSVs in adjacent lane at 0.5m (below the lateral clearance in FTD 227) only in the following exceptions:
    a. AV takes care to reduce its speed relative to the TSV, with an appropriate speed such as 20% higher than the TSV. In case the TSV is stationary in traffic, the AV can pass at a maximum of 30km/h at this reduced clearance of 0.5m
    b. The AV could bias to the left of the lane to maintain 0.5m to TSV in adjacent lane: only when passing the TSV. And only if the vehicle profile extremity (profile at mirror) does not exceed the kerb.

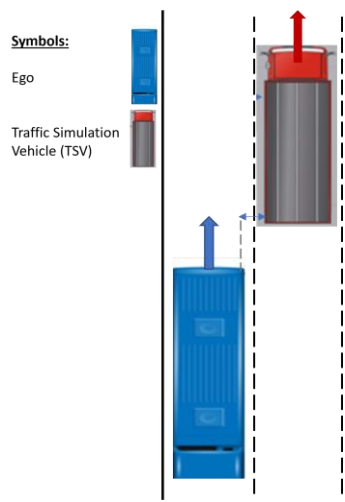
Figure 14: Ego staying behind wide TSV in adjacent lane.

### 4.5.2 Vegetation on left side of lane

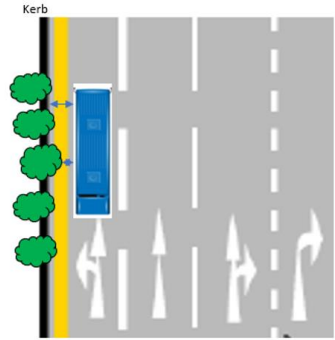

Figure 15: Traffic situation of protruding vegetation from left side of road.

#### *4.5.2.1   General Situation*

**Description**

- Pockets of protruding vegetation along extreme left lane where ego is driving pass.

**Assumptions**

- Ego moving along straight road in lane.
- Multiple lanes on road
- Vegetation within height of bus within some margin.

**Difficulty**

- Human drivers have the capability to estimate that small branches do not cause harm and they would drive through small branches. AV's difficulty is in the perception granularity. The AV could detect vegetation as an object and its decision is to avoid.
    - Could affect trajectory and not comfortable for passengers.
- Driveable area of lane is not defined

**Current guidelines**

- If Lane Change: **TR68-1: 7.6)** Overtaking Outlines all the steps.
- **TR68-1: 7.4b)** The AV shall keep to the left of a two-lane carriageway, except when overtaking **(BTD67).**
- **TR68-1 7.4d)** The AV shall keep within its lane unless it is performing a lane change or overtaking manoeuvre **(FTD311:** Keep within your lane. It is inconsiderate to straddle lanes as this would obstruct other vehicles and may lead to traffic being held up**).**
- **FTD 226**: when passing fixed obstacles: **0.5m**
- **BTD 160c), FTD312** Do not weave in and out of traffic lanes as it would cause confusion and danger to others.
- **TR68-1: 7.5a)** Before commencing a lane change manoeuvre, the AV shall ensure that the manoeuvre does not force any ORU to take potentially unsafe evasive action.

### *4.5.2.2 Analysis of Situation*

This situation was raised by STEAS as with a larger vehicle size in height, they experience such protruding pockets of vegetation frequently on Singapore roads, a common feature on Singapore's tropical road conditions. The issue with current AV technology is that the current state of object classification does not have granularity in differentiate a thick branch or a thin branch and weather the vegetation is passable or not. This would require a bus to slightly weave within lane or execute lane change often to maintain a lateral clearance to protruding vegetation. An earlier recommended guideline to drive within the centre of the lane and defined driveable area within a lane applies to this situation as well. Some other traffic situations discussed with STEAS utilize this same recommended behaviour of straddling lanes and possible lane change decision are included in the appendix section 2 on situations regarding lane change.

The following additional recommended guideline for this situation is:

1. If the lane is wide enough to maintain 0.5m to vegetation and maintain lateral clearance to vehicles (1.5m FTD 227) in the adjacent lane, AV can bias to right of driving lane (within driveable area)
2. If there are no vehicles in adjacent lane when passing the protruding pockets of vegetation
   a) The AV is allowed to exceed driveable area and straddle 2 lanes to maintain 0.5m to vegetation for length below 80metres (between 2 consecutive lamppost) when there are no vehicles in adjacent lane.
   b) If need to straddle lanes for more than 80m, the AV should execute standard full lane change manoeuvre.
3. If there are vehicles in adjacent lane when passing the protruding pockets of vegetation
   a) The AV should slow down and stop if necessary until it is safe to pull into other lane.
   b) When changing lanes, the AV should be maintaining a safe distance of 1 car length per 16km/h of the TSV speed or 9 seconds TTC whichever is bigger to the TSV/cyclist coming from behind in the adjacent lane (i.e. when the AV starts to cross the lane marking)

### 4.5.3 Pedestrian on road

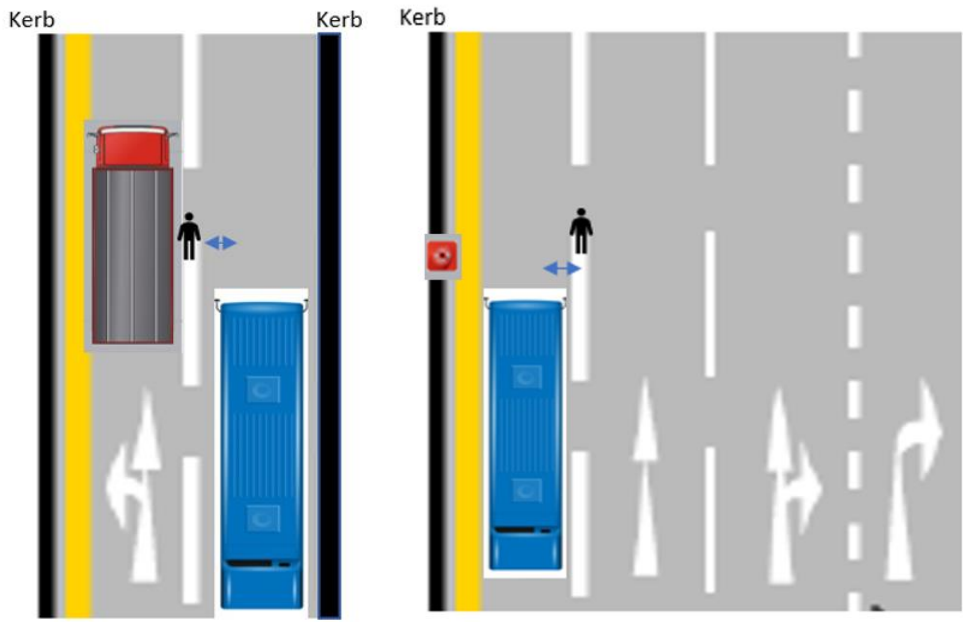

Figure 16: Left: Traffic situation where pedestrian is on road next to large vehicle. Right: Traffic situation of a pedestrian on road surface, with object on road at kerbside

#### *4.5.3.1 General Situation*

**Description**

- Left Situation: Pedestrian standing beside a large vehicle when Ego is driving pass.
- Right Situation: Pedestrian jaywalks halfway across on road and stops on white road marking when Ego is approaching. Cone on road surface at same longitudinal location from AV as pedestrian

**Assumptions**

- Ego driving along straight road
- Left Situation: Road has two lanes; with physical kerb on right of right lane.
- Right Situation: Multiple lanes on road
- Lane width is insufficient / sufficient to maintain lateral clearance to pedestrian.

**Difficulty**

- In these traffic situations, it may render an AV unable to pass through the road to maintain this lateral clearance from pedestrians. **TR68-1: 6.3.2 c) & d)**
- Can the lateral clearance be violated? If so, what should be the minimum lateral clearance allowable to pass the object (pedestrian, car, cone, etc?)
- Is there priority to give based on the lateral clearance? If so, what should be the priority?

**Current Guideline**

- **TR68-1: 6.3.2 c) & d)** when pedestrian is not moving away laterally from AV; and pedestrian on road surface. Pedestrian on road; not moving away
    - AV<30km/h; min 1m
    - AV>30km/h: min 1.5m
- **TR68-1: 6.3.2 e)** when pedestrian is moving away, on road surface
    - Pedestrian on road, moving laterally away: Min 1m
- **FTD226** When passing fixed obstacles, keep a gap of at least 0.5 m from them

### *4.5.3.2 Analysis*

In such a traffic situation of a dual lane road in the left of Figure 16 above, if an AV encounters this traffic situation and has to maintain a 1- 1.5 metres clearance to a pedestrian, there may not be sufficient width on the road for a wide AV like a Class 4 vehicle (about 2.6metre width) to drive past and maintain this lateral clearance and not strike the kerb on the right side. For the traffic situation on the right of Figure 16 above, the AV is also faced with the difficulty of driving through the and maintaining the current 1- 1.5 metres clearance to a pedestrian and 0.5metre to a stationary object in the current lane dimensions. CETRAN and STEAS discussed to recommend a possibility to reduce the AV speed to allow for a lower lateral clearance to pedestrians at 0.5 metre. CETRAN and STEAS conducted a test to check on this and confirmed it was acceptable. Some other traffic situations discussed with STEAS utilize this same recommended behaviour of reducing lateral clearance to pedestrians are included in the appendix section 2 on situations regarding lateral clearance to pedestrians.

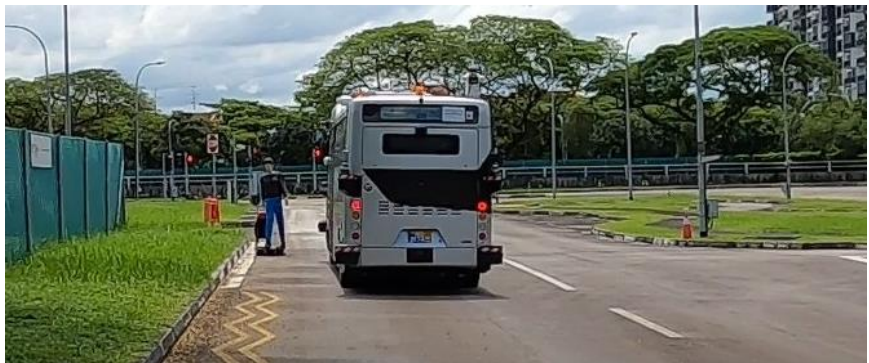

Figure 17: Testing at CETRAN test track for clearance to pedestrian on road with STEAS Class 4 vehicle.

The additional recommended guideline for passing pedestrians on road from this traffic situation analysis are:

1. AV should maintain lane discipline – drive in centre of driveable area in lane if able to keep lateral clearance as per **TR68-1: 6.3.2 c) & d)**
2. It is acceptable for an AV to bias within the lane within the driveable area to maintain required clearance to pedestrian.
3. If unable to maintain 1.5m clearance to a stationary pedestrian above 30km/h; Slow down to below 30 km/h (1m clearance); otherwise if 1m lateral clearance cannot be maintained, a minimum of 0.5m to the pedestrian is allowed if the AV stays below 30km/h as an exception to maintain traffic flow.
4. If unable to maintain 0.5m to pedestrian, slow down and stop to allow pedestrian to move. If an AV is unable to proceed after 10seconds –alert control centre.

# 5. Workshop with AV Developers

A workshop was held with AV developers who have presence in developing in Singapore to share about CETRAN's work in this UMGC project and to gather feedback from the industry of challenges in developing driving behaviour using current guidelines for Singapore road. This workshop was held on 16th January 2023 at a conference room at the Energy Research Institute (ERIAN) at the Cleantech One office. Developers who attended the session were: Moovita, Venti, Desay, STEAS. This provided some diversity in the trial areas and experience.

A worked example of a traffic situation was utilized to go through applicable current guidelines and highlight the current challenges and CETRAN's recommendation for improvement. This aided in targeted discussions with the developers on certain specific guidelines.

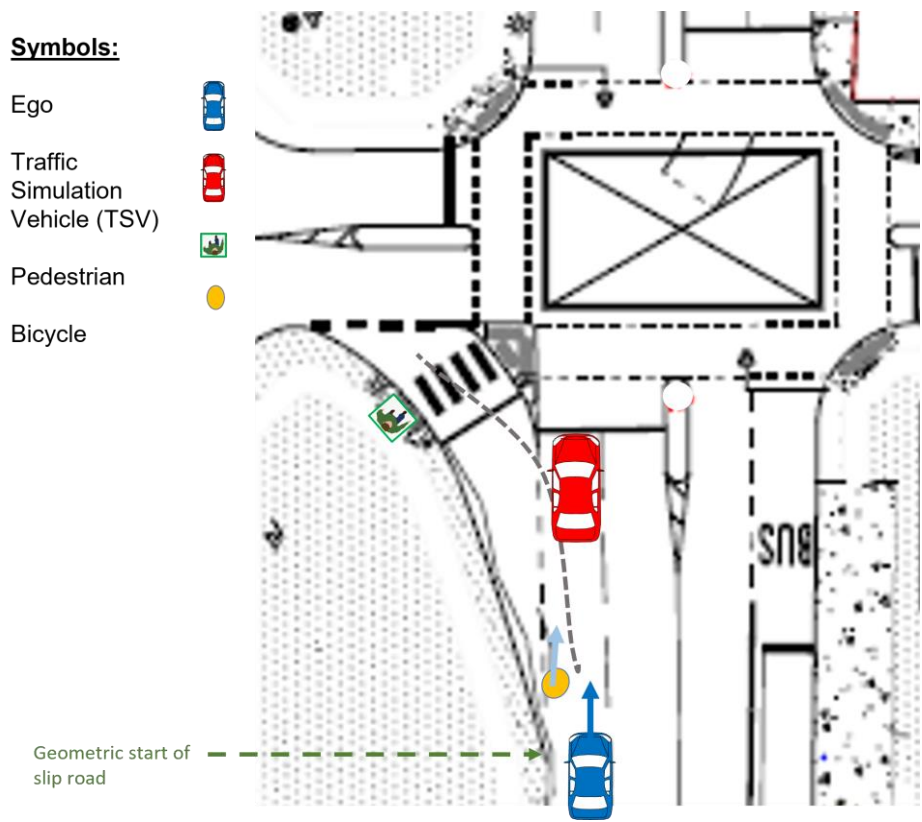

Figure 18: Traffic situation where an AV (ego vehicle) is approaching a left turn slip road with other traffic actors depicted.

The worked example was of a generic situation of approaching a left turn slip road seen in Figure 18 above, with the decision to overtake a bicycle up ahead before the slip road. Within the slip road, the AV would encounter a zebra crossing with potential pedestrians and another traffic simulated vehicle (TSV) stationary at the traffic junction which could cause an occlusion for viewing traffic situation. Prior to the workshop, questions were sent to attendees to allow developers to prepare for the session and share incidences of challenging traffic situation where current guidelines are difficult to follow.

It led to overall discussions on the practicality of applying current driving guidelines for humans on AV, practical metrics should be updated for assessing AVs for road trials in general. The following points are a summary of the feedback received from AV developers active in preparation for road trials in Singapore.

Overall comments:

1) Developers are interested in publicly available situations and specific compliance expectation. Published expected behaviours would help developers with the knowledge on expected competencies in Singapore. Waymo safety report lists specific behavioural competencies such as

    - High speed merge/low speed merge
    - Detect and response to oncoming vehicle/ stopped vehicle

2) Some other standards / guidance developers use include:

    - UNECE R13, R79, R157
    - Bus captain guidelines,
    - Publicly available scenarios (Waymo [12], Carla challenge [13], Safetypool [14])

3) Developers shared their experience of some situations with conflicting rules in Singapore:
    - Overtaking parked vehicle when there is a double white line
    - Maintaining safety (lateral clearance) vs traffic flow (e.g., pedestrian on the road)
    - Road works limits → e.g. Only right turn lane available but AV wants to turn left direction.
    - Following distances (2seconds, 1 car length per 16km/h)

4) Developers shared some instances of missing guidance:
    - <u>Edge cases</u>.
        - Non- compliance from human drivers (taxi illegal drop off at zebra crossing)
        - Vehicle breakdown/ emergency vehicles
    - <u>Common scenarios</u>
        - Junctions without yellow box
        - Bus stops – behaviour at bus stop

5) Difficult situations for AV to manage:
    - Emergency vehicles
    - Traffic conductor hand signals/signage
    - Traffic light breakdown
    - Construction and temporary lane markings
    - Loss of friction difficult to detect e.g. wet road surface
    - Mixed traffic (human/robot) generally is hard for human non-compliance
    - Discretionary right turns
    - Roundabouts.
    - Negotiating narrow lanes.
    - Occlusion – double storey bus
    - Perception items, generally non-behavioural or just emergent behaviour (to slow down)
        - weather, heavy rain
        - plants overhanging (passable/non-passable object)
        - prediction of pedestrians' intention, attentiveness +cyclists
        - animal classification: bird (not to slow) vs chicken (to slow)

Overall, AV developers acknowledge it is difficult to define guidance generically and determining acceptable levels of violation often leads to long debates otherwise. The current AV requirements applies to all situations and could have allowance for variation depending on situations, allowing for practicality and flexibility. An example of this is to have lateral clearance varying based on situations.

AV developers also shared that the current road rules are meant for human driving, and some are difficult to follow. Some suggestions include to propose updated guidelines to consider the intention of the rules and capability of AVs where robots have more precise measurement capability. Situations that could make use of this is in the example of following other vehicles in traffic. Currently a 2-seconds following guide for human driving includes human reaction time, is an easy method to apply but is very conservative and not often used by drivers on the road. One suggestion would be to have a 2-metre clearance upon stopping, to be calculated by AV based on its own braking capability. Another method raised was to follow Mobileye's Responsible Sensitive Safety (RSS) [15] method to calculate the required distance.

# 6. Conclusion

## 6.1 Developers Workshop outcome

Through our consultation with AV developers on improving guidelines for AV behaviours under this UMGC project, AV developers acknowledge it is difficult to define guidance generically and determining acceptable levels of violation often leads to long debates otherwise. The current AV requirements applies to all situations and could have allowance for variation depending on situations, allowing for practicality and flexibility. Some suggestions from the industry were to propose updated guidelines to consider the intention of the rules and capability of AVs where robots have more precise measurement capability. AV developers also had a request to have publicly published scenarios and corresponding expected behaviours that will help in developing AV behaviour for Singapore roads.

## 6.2 Recommendations from project

From this study, CETRAN has identified some recommendations for additional guidelines from selected traffic situations chosen from testing experience and from the first consolidation of existing guidelines in Singapore. A summary of these additional recommended guidelines is listed below. Insights gathered earlier should be noted from the workshop conducted with AV developers (seen in Section 5) for further improvements to defining acceptable AV behaviour for Singapore roads. These recommendations can be sent to TR68-1 working committee as considerations to update the standards for AV behaviour in Singapore.

Summary of additional recommended guidelines:

1) Utilize a safe following distance of "so that if the leading vehicle would initiate a complete stop, the AV would be able to stop behind that vehicle with a longitudinal distance of at least 2 m" in TR68-1 6.3.4 to be used for assessment.
   **[Manoeuvres – Driving in Lane with Traffic – Following Actor]**
2) Complete overtaking manoeuvre with 3 seconds before start of left turn slip road
   **[Manoeuvres – Overtaking; Traffic Infrastructure – Signalized Junction -Turns- Left turn slip road]**
3) Safe distance for lane change to be: **[Manoeuvres – Ego Lane Change]**
   a) If there is an oncoming vehicle in the adjacent during the overtaking manoeuvre, the AV should change back into its original lane with at least a 2 second time gap (time to collision) with the TSV/cyclist oncoming in the adjacent lane.
   b) When changing lanes, the AV should be maintaining a safe distance of 1 car length per 16km/h of the TSV speed or 9 seconds TTC whichever is bigger to the TSV/cyclist coming from behind in the adjacent lane (i.e. when the AV starts to cross the lane marking)
4) A recommended value for lateral acceleration should be used in TR68-1, such as UNECE 79 5.6.2.1.3 [8] according to vehicle class and vehicle speeds, such that bad behaviour of 'sudden cut ins' can be better defined, together with safe distances to other actors. This is recommended to follow a maximum of $3m/s^2$ for speeds between 10-60km/h and with a moving average over half a second of the lateral jerk generated to not exceed $5m/s^3$ [8].
   **[Manoeuvres – Ego Lane Change]**
5) When entering a left turn slip road with a zebra crossing approaching a traffic junction: An AV should start turning at 3 metre after the start of the geometric start of slip road, maintaining a smooth trajectory from the centre of the original lane into the centre of the left turn slip road lane, so as not to encourage motorcyclists from entering from the left of the AV. (This is applicable for urban roads and not for exits from highways). **[Manoeuvres – Overtaking; Traffic Infrastructure – Signalized Junction -Turns- Left turn slip road]**

6) If an AV vehicle length is longer than the available space between a zebra crossing and the give-way line, the vehicle should wait at the stop line before a zebra crossing, until it can safely join traffic coming from the right. If however, an AV's view of oncoming traffic is occluded when stopped at the zebra crossing, it can creep forward when there are no pedestrians crowing and block the zebra crossing until it is safe to complete the left turn. **[Manoeuvres – Overtaking; Traffic Infrastructure – Signalized Junction -Turns- Left turn slip road]**

7) If there is overlap to AV's next action-another right turn within 10s, continue signal, otherwise cancel signal. **[Manoeuvres – Ego Lane Change]**

8) An AV should cross the path of an oncoming TSV/cyclist in a discretionary right turn if it is able to complete the crossing with at least a 2 second time buffer to the oncoming TSV/cyclist at the end of the turn, out of the traffic junction. **[Traffic Infrastructure – Signalized Junction -Turns- Discretionary right turn]**

9) Applying the lateral clearances necessary from TR68-1 7.9.4 from zebra crossing to be applied to a pedestrian crossing at a traffic junction as well. **[Traffic Infrastructure – Pedestrian Crossing]**

10) If an AV is in the left lane of the 2 lanes turning, it should turn into the 2$^{nd}$ lane from the left of the central divider (e.g. middle lane of 3 lanes). However, if there is a guiding lane that directs traffic differently, the guiding line should be prioritized. **[Traffic Infrastructure – Signalized Junction -Turns- Turning Right; Traffic Infrastructure – Unsignalized Junction -STOP Line- Right Turn; Traffic Infrastructure – Unsignalized Junction -Turns from Major road- Right Turn]**

11) An AV should show caution when overtaking a parked vehicle. To proceed with a lateral clearance of 1metre, it should slow down and pass with a maximum speed of 30km/h to the parked vehicle. **[Manoeuvres – Overtaking]**

12) An AV should return to its original lane if it is clear of traffic, after overtaking the parked vehicle after it has at least 1- 1.5x ego vehicle's car length ahead of the parked vehicle and with a maximum of 20 seconds past the parked vehicle in the other lane. **[Manoeuvres – Overtaking]**

13) Further guidance on acceptable behaviour for buses at bus stop infrastructure, such as lateral distances to pedestrians **[Traffic Infrastructure – Bus stops]**

14) AV should drive in the centre of the drivable area within a lane (as seen in Figure 12 above) **[Lane Discipline-General]**

15) An AV can stay behind a slow TSV in the adjacent lane if the lateral clearance is less than the 1.5m in FTD 227. **[Lane Discipline-General]**

16) The AV is allowed to drive pass the slow moving TSVs in adjacent lane at 0.5m (below the lateral clearance in FTD 227) only in the following exceptions:
    a) AV takes care to reduce its speed relative to the TSV, with an appropriate speed such as 20% higher than the TSV. In case the TSV is stationary in traffic, the AV can pass at a maximum of 30km/h at this reduced clearance of 0.5m
    b) The AV could bias to the left of the lane to maintain 0.5m to TSV in adjacent lane: only when passing the TSV. And only if the vehicle profile extremity (profile at mirror) does not exceed the kerb.

17) If the lane is wide enough to maintain 0.5m to vegetation and maintain lateral clearance to vehicles (1.5m FTD 227) in the adjacent lane, AV can bias to right of driving lane (within driveable area) **[Lane Discipline-General]**

18) If there is no vehicles in adjacent lane when an AV requires to exceed the lane's drivable area (such as passing the protruding pockets of vegetation or static objects or facing a narrow lane during a bend) **[Lane Discipline-General]**
    a) The AV is allowed to exceed driveable area and straddle 2 lanes to maintain 0.5m to vegetation for length below 80metres when there are no vehicles in adjacent lane.

      b) If the need to straddle lanes exceeds 80m, the AV should execute standard full lane change manoeuvre.
19) If there are vehicles in adjacent lane when an AV requires to exceed the lane's drivable area (such as passing the protruding pockets of vegetation or static objects or facing a narrow lane during a bend) **[Lane Discipline-General; Manoeuvres – Ego Lane Change]**
      a) The AV should slow down and stop if necessary until it is safe to pull into other lane.
      b) When changing lanes, the AV should be maintaining a safe distance of 1 car length per 16km/h of the TSV speed or 9 seconds TTC whichever is bigger to the TSV coming from behind in the adjacent lane (i.e. when the AV starts to cross the lane marking)
20) It is acceptable for an AV to bias within the lane within the driveable area to maintain required clearance to pedestrian **[Lane Discipline-General]**
21) If unable to maintain 1.5m clearance to a stationary pedestrian above 30km/h; Slow down to below 30 km/h (1m clearance); otherwise, if 1m lateral clearance cannot be maintained, a minimum of 0.5m to the pedestrian is allowed if the AV stays below 30km/h as an exception to maintain traffic flow. **[Manoeuvres – Driving in Lane with Traffic – Stationary Actor; Manoeuvres – Overtaking]**
22) If unable to maintain 0.5m to pedestrian, slow down and stop to allow pedestrian to move. If an AV is unable to proceed after 10seconds - (alert control centre) **[Manoeuvres – Driving in Lane with Traffic – Stationary Actor; Manoeuvres – Overtaking]**

## 6.3 Future Work

During this study on acceptable AV behaviours, CETRAN has consolidated existing guidelines and identified some conflicting guidance and missing guidance. However further detailed analysis using identified traffic situations allowed another avenue for investigation into further missing or conflicting guidance from current guidelines. Collaboration with STEAS allowed for additional insights into further traffic situations encountered by a developer with missing guidance. Some of these situations were selected for further testing on CETRAN test track to consolidate and evaluate recommended additional guidelines. The results of these discussions with STEAS on selected situations from STEAS experience and tests on relevant parameters will be summarized in a joint white paper by CETRAN and STEAS. These allowed for a first insight to the potential of improvement of guidelines for AV behaviour in Singapore's urban traffic environment. Improving this would provide developer companies a concise and clear expectations of AV behaviour when exploring Singapore as a potential AV trial environment. Consultation with AV stakeholders such as AV developers could be conducted in future projects.

Further discussions could be held with Traffic Police to understand some commonality of improving recommended guidelines. TP is conducting a trial on Intelligent Driving Circuit (IDC) at Singapore Safety Driving Centre [16]. CETRAN can share these recommendations on improved guidelines for AV behaviour to TP.

The limited situations were selected based on priority and importance; further work could be done to extend to other situations for an overall comprehensive evaluation on acceptable AV behaviours for trials on Singapore roads. Further robust testing needs to be carried out to recommend with increased confidence the additional guidelines CETRAN has listed above. Some of these recommendations were tested at limited speeds due to the road length constraints of the CETRAN circuit track, which may not apply to the wide range of road situations. Further situations, realistic road speeds would require robust testing to be carried out.

# 7. Acknowledgements

We acknowledge our current partners who supported us in this project. In particular, we thank the team from Singapore Technologies Engineering Autonomous Solutions (STEAS) for their valuable support sharing insights and experiences regarding real challenges faced by Autonomous Vehicles driving in Singapore roads. Additionally, for providing the use of their AV bus and specialized staff to verify the practicality of the recommended guidelines for AV expected behaviour at CETRAN test track.

We also thank all the AV developers, including Moovita, Venti, Desay and STEAS, who attended the workshop session we organized for their constructive feedback and support.

This research/project is supported by the National Research Foundation, Singapore, and Land Transport Authority (Urban Mobility Grand Challenge (UMGC-L010)). Any opinions, findings and conclusions or recommendations expressed in this material are those of the author(s) and do not reflect the views of National Research Foundation, Singapore, and Land Transport Authority.

# 8. Acronyms

The terms and abbreviations used in the document are listed below.

| | |
|---|---|
| AASHTO | American Association of State Highway and Transportation Officials |
| AV | Autonomous Vehicle |
| BTD | Basic Theory of Driving |
| CETRAN | Centre of Excellence for Testing & Research of Autonomous Vehicle |
| FTD | Final Theory of Driving |
| ISO | International Organization for Standardization |
| LTA | Land Transport Authority |
| ORU | Other Road User |
| SAE | Society of Automotive Engineers |
| TR68-1 | Technical Reference 68 Part 1 |
| TSV | Traffic Simulated Vehicle |
| TTC | Time to Collision |
| UNECE | The United Nation Economics Commission for Europe |
| VRU | Vulnerable Road User |
| VUT | Vehicle Under Test |
| WP | Work Package |

# 10. Appendices

## A. Calculation of braking distance

An initial calculation for following distance was done using the American Association of State Highway and Transportation Officials (AASHTO) from the formula shown in Figure 19 below. This was then compared to current BTD guidelines for following distance such as 2 seconds following and 1 car length per 16km/h speed. It appears that the braking distance from AASHTO with an additional 2metres when stopped follows the guideline of 1 car length per 16km/h at low speeds before diverging at higher speeds. This could incorporate that at higher speeds, a vehicle's braking distance is non-linear to its travelling speed, differing from the guideline of the linear 1 car length per 16km/h. It should be noted that an approximate braking friction coefficient of 0.6 was used, and what this implies for braking deceleration is not known. Hence the recommendation that developers should use the braking distance with the known vehicle dynamics would be a better estimate and allows for various developers to incorporate their individual vehicle characteristics. The guideline of 2seconds following or 1 car length per 16km/h does not incorporate such elements and may not apply uniformly across various vehicle sizes.

The AASHTO formula is as follows:

$$s = (0.278 * t * v) + v^2 / (254 * (f + G))$$

where:

- **s** is the stopping distance, measured in meters;
- **t** is the perception-reaction time in seconds;
- **v** is the speed of the car in km/h;
- **G** is the grade (slope) of the road, expressed as a decimal. It is positive for an uphill grade and negative for a road going downhill;
- **f** is the coefficient of friction between the tires and the road. It is typically assumed to be equal to 0.7 on a dry road and in the range from 0.3 to 0.4 on a wet road.

Figure 19: AASHTO braking distance formula.

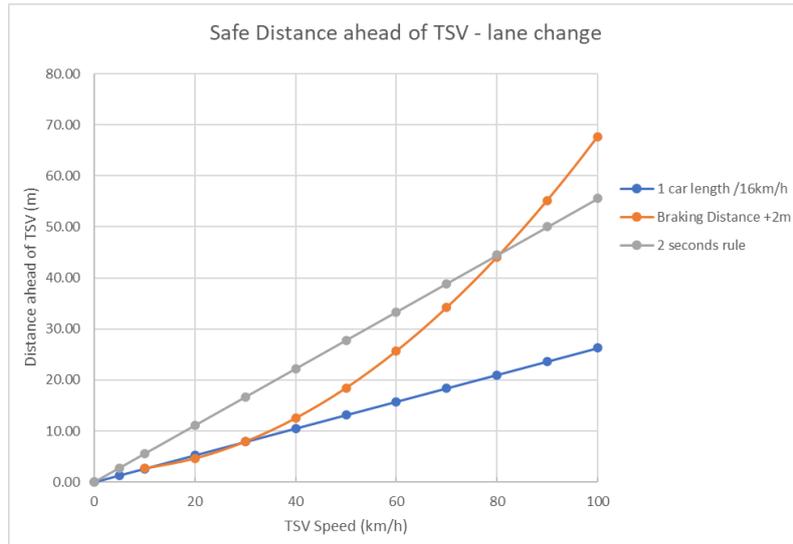

Figure 20: Graph of braking distance calculated using AASHTO formula with an approximated friction coefficient of 0.6.

## B.   Other traffic situations discussed with STEAS

These situations have been discussed with STEAS, and the corresponding recommended behaviour is represented in the examples in Section 4.5.

**Situations regarding lane discipline**

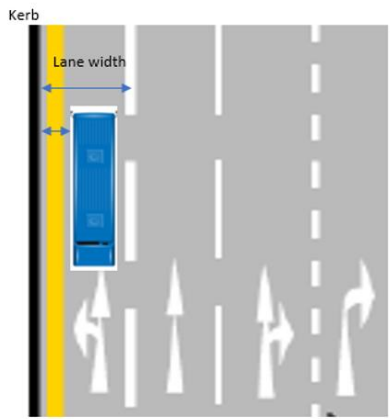

Figure 21: Drivable area in lane

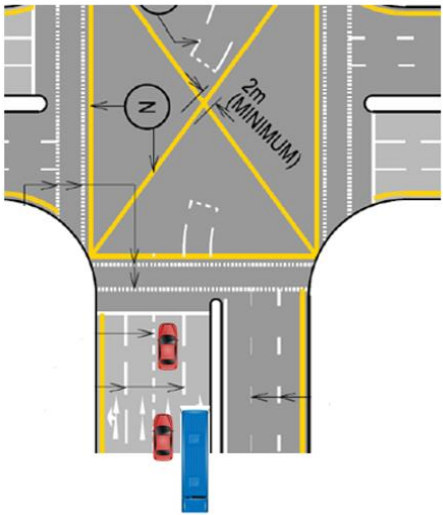

Figure 22: Bus pulling into narrow right turn lane.

**Situations regarding lane change**

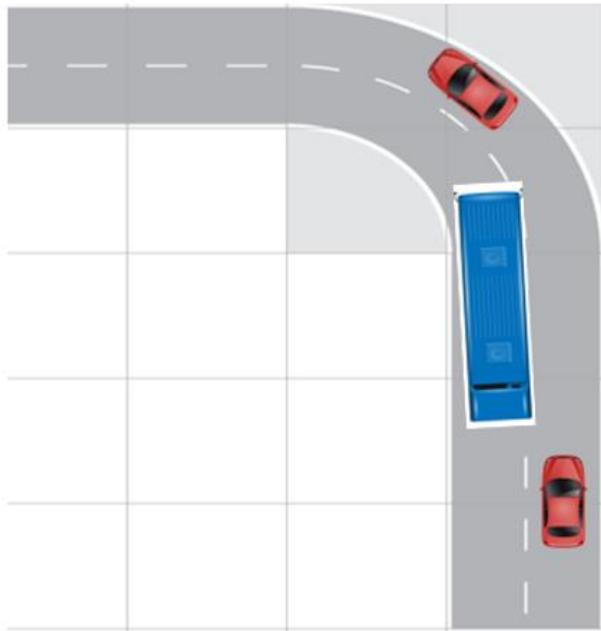

Figure 23: Lane keeping in bend - requiring lane change?

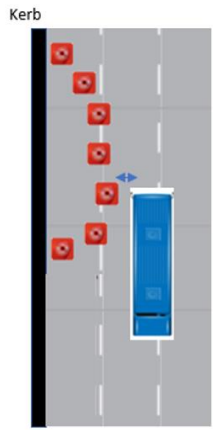

Figure 24: Cones on road, requiring lane change?

**Situations regarding lateral clearance to pedestrians**

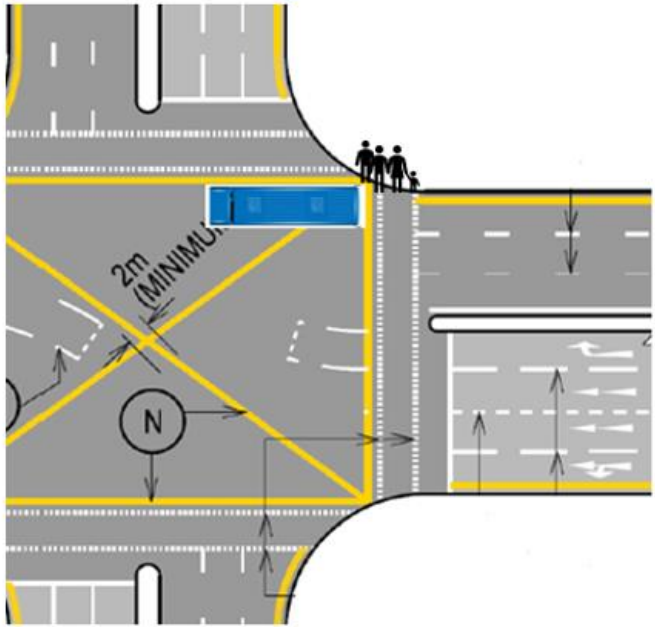

Figure 25: Pedestrian on kerb at junction

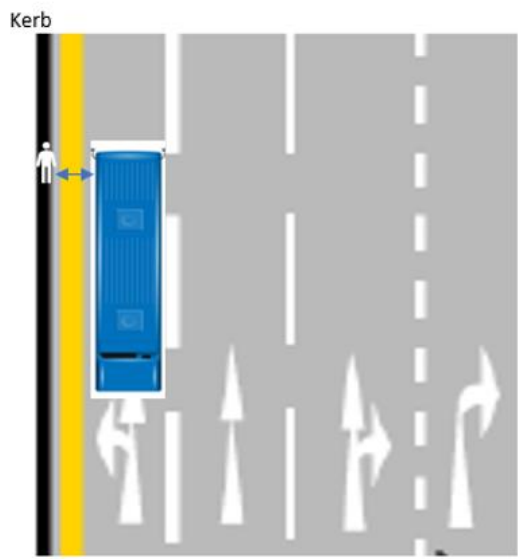

Figure 26: Pedestrian on road at kerbside